\documentclass[twoside]{article}

\usepackage[preprint]{lib/aistats2027}
\usepackage[round]{natbib}

\newcommand{\Norm}{\mathcal{N}}
\newcommand{\sm}{\mathrm{softmax}}
\newcommand{\bdelta}{\bm{\delta}}
\newcommand{\btheta}{\bm{\theta}}
\newcommand{\bmu}{\bm{\mu}}
\newcommand{\brho}{\bm{\rho}}
\newcommand{\balpha}{\bm{\alpha}}

\usepackage{amsmath,amsfonts,bm}

\def\eqref#1{equation~\ref{#1}}

\def\1{\bm{1}}

\DeclareMathAlphabet{\mathsfit}{\encodingdefault}{\sfdefault}{m}{sl}
\SetMathAlphabet{\mathsfit}{bold}{\encodingdefault}{\sfdefault}{bx}{n}

\newcommand{\KL}{D_{\mathrm{KL}}}

\newcommand{\sentetm}{SentenceETM}
\usepackage{pifont}

\usepackage{autobreak}
\usepackage{booktabs}
\usepackage{soul}
\usepackage{subcaption}
\usepackage{graphicx}
\usepackage{multirow}
\usepackage{tabularx}
\usepackage{hyperref}
\usepackage{url}
\usepackage{cleveref}
\usepackage{xcolor}
\usepackage{algorithm}
\usepackage{algpseudocode}
\usepackage{wrapfig}
\definecolor{sportbg}{HTML}{C8DCF5}
\definecolor{techbg}{HTML}{F7D6C9}

\newcommand{\sentA}[1]{{\sethlcolor{sportbg}\hl{#1}}}
\newcommand{\sentB}[1]{{\sethlcolor{techbg}\hl{#1}}}

\newcommand{\Mult}{\operatorname{Mult}}

\newcommand{\bsigma}{\bm{\sigma}}
\newcommand{\bSigma}{\bm{\Sigma}}
\newcommand{\blambda}{\bm{\lambda}}

\begin{document}

%
\runningtitle{SentenceETM}

%

\twocolumn[

\aistatstitle{Sentence-Level Topic Modeling\\
with \textit{Lexical} and \textit{Contextual} Evidence}

\aistatsauthor{ Federica Granese \And Serena Villata \And  Charles Bouveyron }

\aistatsaddress{ Université Côte d’Azur,\\CNRS, I3S, Inria \And  Université Côte d’Azur,\\CNRS, I3S, Inria \And Université Côte d’Azur,\\CNRS, LJAD, Inria } ]

\begin{abstract}
Topic modeling faces a fundamental granularity dilemma: probabilistic topic models assign topics at the word level, whereas transformer-based methods operate at the document level. The latter overlooks that real-world documents typically cover multiple topics, whereas the former relies on individual words, which rarely provide sufficient context to identify them. We introduce SentenceETM, a neural topic model that assigns a topic to each sentence by jointly using lexical and contextual evidence. Each topic is described both by a distribution over the vocabulary, defined through word embeddings, and by a Gaussian over sentence embeddings concentrated on a topic-specific low-dimensional subspace. The topic of a sentence generates both its words and its embedding and is marginalized exactly, so that its posterior combines the two sources of evidence. Document proportions are inferred with amortized variational inference, while the Gaussian parameters are re-estimated in closed form. 
Experiments show that SentenceETM consistently improves sentence-level topic assignment over state-of-the-art approaches, and a case study of a presidential debate shows how sentence-level topics reveal how candidates handle the same subjects.
\end{abstract}

\section{Introduction}
\label{sec:intro}
Topic models are well-established methods for extracting the main themes from a collection of documents. Regardless of their underlying formulation, these methods produce a set of topics that characterize the corpus, where a topic is broadly understood as a group of semantically coherent words. A document is characterized by a mixture of these topics. However, such a mixture tells us \textit{which} topics the documents contain but not \textit{where} they are localized. Indeed, real documents rarely contain a single theme, and locating them requires deciding which textual unit a topic is attached to and from which \textit{evidence} it is inferred.  

We refer to \textit{lexical evidence} as the information carried by words occurring in a text segment\footnote{We use the term `text segment' to refer generally either to a document in its entirety or to a part of it.}, represented as counts over a fixed vocabulary, and to \textit{contextual evidence} as the dense representation of that text segment produced by a pretrained neural encoder. Lexical evidence is directly interpretable as a distribution over words, but it is blind to word order and to the sense a word takes in a given occurrence. Contextual evidence can capture the overall semantics of a document more faithfully, but at the cost of interpretability, since the words that characterize a topic are no longer part of its representation and must be recovered separately.

The two main families of topic models each build primarily on one of these types of evidence, because the textual unit they adopt admits primarily one of them.
Bag-of-words approaches, e.g.~\citep{blei2003latent}, treat the word as the fundamental unit: each word receives a latent topic assignment, and a document becomes a mixture over topics. A word is represented independently of its surrounding context, so contextual evidence from the document is not available to the model at this level.
Embedding-based approaches, e.g.~\citep{grootendorst2022bertopic}, take the opposite stance, treating the document as the fundamental unit and clustering the encoder's representations. Contextual evidence is exploited directly, but lexical evidence re-enters only post hoc, through a descriptor-extraction step playing no role in the topic assignment itself. The result is that one family is too fine-grained to use context, the other too coarse-grained to localize the topics.

Sentence-level topic models lie in between these two extremes. A sentence is semantically self-contained, short enough that its words and its embedding are concerned with a common theme, and numerous enough within a document that assigning a topic to each sentence reveals how topics are distributed throughout the text. Most importantly, at the sentence level, lexical and contextual evidence became available at the same textual unit. While sentence-level topic models have been proposed in the past~\citep{balikas2016topic,cha-lee-2024-sentencelda}, each model still relies on a single type of evidence for topic assignment.
To fill this gap, we make the following contributions.

(i) We introduce {\sentetm}, a neural sentence-level topic model that jointly uses lexical and contextual evidence to assign a topic to each sentence. Each topic is characterized by both a distribution over the vocabulary, defined via word embeddings, and a region of the sentence-embedding space, modeled as a Gaussian concentrated on a topic-specific low-dimensional subspace, thereby keeping the model tractable in the high-dimensional space of sentence encoders. Each sentence draws a topic from the document's topic proportions, and that topic generates both the sentence's words and its contextual embedding. The sentence topic is marginalized exactly, so that its posterior combines the two sources of evidence, and the parameters of the contextual side are estimated in closed form.

(ii) We present a novel evaluation pipeline for sentence-level topic models. We build synthetic corpora from existing topic-labeled document collections, in which every sentence has a known topic and document-level topic mixtures are controlled by construction.

(iii) We show that {\sentetm} can be adapted to different domains through an interchangeable sentence encoder, without modifying the generative model. We illustrate this on the transcript of a televised presidential debate. The sentence-level representation allows us to examine how candidates distribute topics within the same debate segment and how frequently new topics are introduced during speaking turns.
\subsection{Related works}
\label{sec:related}
The use of sentences as a textual unit in topic modeling has been previously investigated mainly as an extension of LDA~\citep{blei2003latent}. Early approaches incorporated sentence- or segment-level structure without making the segment the unit of topic assignment~\citep{boyd2008syntactic,wang2009multi,jiang2019sentence}. Differently, senLDA~\citep{balikas2016topic} moves the topic assignment from words to sentences: one topic is drawn for each sentence, and the words in that sentence are drawn from the selected topic. The topic--word association is preserved, with each topic represented as a distribution over the vocabulary. More recently, \citet{cha-lee-2024-sentencelda} introduced SentenceLDA, which replaces the word with the sentence as the basic unit of Gaussian LDA~\citep{das2015gaussian}, while \citet{schneider2023efficient} uses sentence embeddings for topic assignment through hard expectation maximization. 
In these methods, contextual information is used directly for topic assignment, whereas lexical information is used only to describe the resulting topics. Our contextual component is related to model-based clustering in high dimensions~\citep {bouveyron2007high}: instead of the full-covariance Gaussians of Gaussian LDA, each topic is modeled by a Gaussian concentrated on its own low-dimensional subspace, which remains tractable for sentence embeddings with hundreds of dimensions.
A parallel line of work combines lexical and contextual evidence without changing the textual unit. Contextualized Topic Models~\citep{bianchi2021pre,bianchi2021cross} combine bag-of-words representations with pretrained contextual document embeddings, while TNTM~\citep{reuter2025probabilistic} models topics as distributions over transformer representations. Both combine the two forms of evidence at the document level and return a topic per document.

\section{The proposed approach}
\label{sec:method}
\begin{figure}[!t]
    \centering
    \includegraphics[width=\columnwidth]{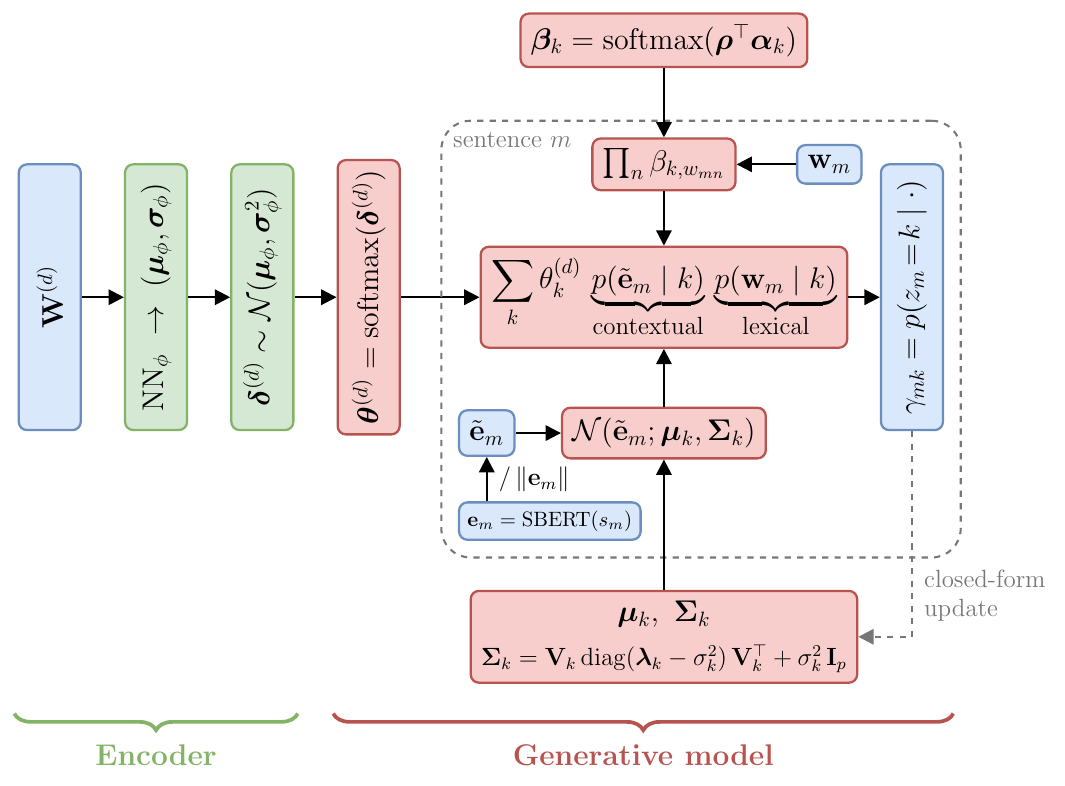}
    \caption{\textbf{Overview of {\sentetm}}. 
    Document-level topic proportions (green) are combined with the contextual and lexical evidence of each sentence (red) into the posterior $\gamma_{mk}$, which re-estimates each topic's subspace Gaussian in closed form (dashed arrow).
    }
    \label{fig:setm_architecture}
\end{figure}

{\sentetm} (\Cref{fig:setm_architecture}) is built on a single question:
 
``\textit{How likely is it to observe a given sentence embedding together with its words, given that the sentence belongs to a document with topic proportions $\btheta^{(d)}$?}''
 
Let $\mathbf w_m^{(d)}=(w_{m1}^{(d)},\dots,w_{mN_m^{(d)}}^{(d)})$ denote the words of the $m$-th sentence of document $d$ and $\tilde{\mathbf e}_m^{(d)}$ its normalized sentence embedding. Given $\btheta^{(d)}$, the model assigns to each sentence the likelihood
\begin{align}
\begin{aligned}
&p\big(\mathbf w_m^{(d)},\tilde{\mathbf e}_m^{(d)}\mid\btheta^{(d)}\big)=\\
&\sum_{k=1}^{K}
\theta_k^{(d)}\underbrace{\Norm\big(\tilde{\mathbf e}_m^{(d)};\bmu_k,\bSigma_k\big)}_{\text{Contextual Evidence}} \underbrace{\prod_{n=1}^{N_m^{(d)}}\beta_{k,w_{mn}^{(d)}}}_{\text{Lexical Evidence}}.
\end{aligned}
\label{eq:sentence_likelihood}
\end{align}
Each term of the sum is the product of the prevalence of topic $k$ in the document, the likelihood of the sentence embedding under the topic, and the likelihood of the sentence words under the topic. Since both views (i.e., evidence) are generated by the same topic, a topic contributes to~\Cref{eq:sentence_likelihood} only when it explains the embedding \textit{and} the words simultaneously. Normalizing the terms of the sum yields the posterior over the topic of the sentence,
$\gamma_{mk}^{(d)}=p\big(z_m^{(d)}=k\mid\mathbf w_m^{(d)},\tilde{\mathbf e}_m^{(d)},\btheta^{(d)}\big)
\propto\theta_k^{(d)}\,\Norm\big(\tilde{\mathbf e}_m^{(d)};\bmu_k,\bSigma_k\big)\prod_{n=1}^{N_m^{(d)}}\beta_{k,w_{mn}^{(d)}}$, which provides the sentence-level topic assignment. 
In practice, the lexical and contextual terms can have different numerical scales; during training and inference, we weight the contextual log-likelihood by $c>0$, with $c=1$ recovering~\Cref{eq:sentence_likelihood}.

\subsection{Generative model}
\label{sub_sec:generative}
A corpus consists of $D$ documents. Document $d$ is a collection of $M^{(d)}$ sentences, and sentence $m$ has the words $\mathbf w_m^{(d)}$, with $w_{mn}^{(d)}\in\{1,\dots,V\}$. Each sentence is observed through two views. The \textit{lexical view} is its bag of words, while the \textit{contextual view} is the embedding $\mathbf e_m^{(d)}\in\mathbb R^{E}$ of the raw sentence, produced by a frozen sentence encoder. Since sentence encoders are trained to compare sentences by cosine similarity, we $\ell_2$-normalize these embeddings, $\tilde{\mathbf e}_m^{(d)}=\mathbf e_m^{(d)}/\|\mathbf e_m^{(d)}\|$, so that Euclidean distances between them are a monotone function of their cosine similarity. Finally, $\brho\in\mathbb R^{p\times V}$ are pre-trained word embeddings.
 
\textbf{Lexical view.} Each topic $k\in\{1,\dots,K\}$ has a topic embedding $\balpha_k\in\mathbb R^{p}$ in the word embedding space, which defines a distribution over the vocabulary, $\bm\beta_k=\sm(\brho^\top\balpha_k)$, as in ETM~\citep{dieng2020topic}.
 
\textbf{Contextual view.} Each topic is also associated with a region of the sentence embedding space, described by a Gaussian distribution with mean $\bmu_k\in\mathbb R^{E}$ and covariance $\bSigma_k$. Sentence embeddings are high-dimensional, and a full covariance matrix would require $O(E^2)$ parameters per topic. Following~\citep{bouveyron2007high}, we instead adopt a flexible and parsimonious modeling which assumes that the sentences of each topic vary mainly within a topic-specific subspace of dimension $h_k\ll E$, spanned by the orthonormal columns of $\mathbf V_k\in\mathbb R^{E\times h_k}$, and that the remaining variation is isotropic noise:
\begin{align}
\bSigma_k=\mathbf V_k\bm\Lambda_k\mathbf V_k^\top+\sigma_k^2\,\mathbf I_E,
\label{eq:hddc_covariance}
\end{align}
where $\bm\Lambda_k=\operatorname{diag}\big(\lambda_{k1}-\sigma_k^2,\dots,\lambda_{kh_k}-\sigma_k^2\big)$. Along the $h_k$ directions of its subspace, topic $k$ has variances $\blambda_k=(\lambda_{k1},\dots,\lambda_{kh_k})$, with $\lambda_{k1}\ge\dots\ge\lambda_{kh_k}$; in every other direction, it has variance $\sigma_k^2<\lambda_{kh_k}$. A sentence is thus likely under a topic when it lies close to the topic subspace, even if it is far from its mean along the directions in which the topic varies. Since each topic has its own subspace, different topics can vary along different directions of the embedding space. The covariance in~\Cref{eq:hddc_covariance} requires $O(Eh_k)$ parameters per topic, and both its inverse and its determinant have closed forms that never involve $E\times E$ matrices (\Cref{app:hddc}). We collect the model parameters in $\Theta=\{\Theta_{\text{lex}},\Theta_{\text{ctx}}\}$, with $\Theta_{\text{lex}}=\{\balpha_k\}_{k=1}^K$ and $\Theta_{\text{ctx}}=\{\bmu_k,\mathbf V_k,\blambda_k,\sigma_k^2\}_{k=1}^K$.
 
\textbf{Generative process.}
For each document, topic proportions $\btheta^{(d)}=\sm(\bdelta^{(d)})$ are drawn from a logistic-normal prior, with $\bdelta^{(d)}\sim\Norm(0,I_K)$. Each sentence then receives a single topic $z_m^{(d)}\sim\Mult(1;\btheta^{(d)})$, which generates both of its views: the normalized embedding, $\tilde{\mathbf e}_m^{(d)}\mid z_m^{(d)}=k\sim\Norm(\bmu_k,\bSigma_k)$, and every word of the sentence, $w_{mn}^{(d)}\mid z_m^{(d)}=k\sim\Mult(1;\bm\beta_k)$. The full process is summarized in~\Cref{alg:generative_process} (\Cref{app:generative}).
Given the topic of a sentence, the two views are assumed to be conditionally independent:
$p\big(\mathbf w_m^{(d)},\tilde{\mathbf e}_m^{(d)}\mid z_m^{(d)}=k\big)
=\Norm\big(\tilde{\mathbf e}_m^{(d)};\bmu_k,\bSigma_k\big)\prod_{n=1}^{N_m^{(d)}}\beta_{k,w_{mn}^{(d)}}$.
Marginalizing the topic of the sentence with respect to the document proportions yields the sentence likelihood of~\Cref{eq:sentence_likelihood}. Since sentences are conditionally independent given $\btheta^{(d)}$, the joint distribution of a document and its latent proportions is
$p\big(\mathbf W^{(d)},\tilde{\mathbf E}^{(d)},\bdelta^{(d)}\mid\Theta\big)
=\Norm\big(\bdelta^{(d)};0,I_K\big)\prod_{m=1}^{M^{(d)}}p\big(\mathbf w_m^{(d)},\tilde{\mathbf e}_m^{(d)}\mid\btheta^{(d)}\big)$, where $\btheta^{(d)}=\sm\big(\bdelta^{(d)}\big)$. The sentence topics $z_m^{(d)}$ are summed out exactly within each sentence likelihood; the only latent variable is the document-level $\bdelta^{(d)}$, whose posterior we approximate in~\Cref{sub_sec:inference}.
 
\subsection{Inference model}
\label{sub_sec:inference}
The marginal likelihood $p\big(\mathbf W^{(d)},\tilde{\mathbf E}^{(d)}\mid\Theta\big)$ of document $d$ is
\begin{align}
\int\Norm\big(\bdelta^{(d)};0,I_K\big)\prod_{m=1}^{M^{(d)}}p\big(\mathbf w_m^{(d)},\tilde{\mathbf e}_m^{(d)}\mid\btheta^{(d)}\big)\,d\bdelta^{(d)},
\label{eq:marginal}
\end{align}
which is intractable because of the logistic-normal prior. We therefore approximate the posterior of $\bdelta^{(d)}$ with variational inference, and fit the model parameters $\Theta$ by maximizing a lower bound on~\Cref{eq:marginal}. For any distribution $q(\bdelta^{(d)})$, the log marginal likelihood decomposes as
$\log p\big(\mathbf W^{(d)},\tilde{\mathbf E}^{(d)}\mid\Theta\big)
=\mathcal L(q;\Theta)
+\KL\big(q(\bdelta^{(d)})\,\|\,
p(\bdelta^{(d)}\mid\mathbf W^{(d)},\tilde{\mathbf E}^{(d)},\Theta)\big)$,
where $\mathcal L(q;\Theta)=\mathbb E_q\big[\log p(\mathbf W^{(d)},\tilde{\mathbf E}^{(d)},\bdelta^{(d)}\mid\Theta)-\log q(\bdelta^{(d)})\big]$ (proof in Appendix~\ref{app:proof1}). Since the KL divergence is non-negative, $\mathcal L$ is a lower bound (ELBO) of the log marginal likelihood, and maximizing it with respect to $q$ brings $q$ closer to the true posterior.
 
\textbf{Variational assumptions.}
We restrict $q$ to Gaussians with diagonal covariance, amortized by an inference network, to approximate the intractable posterior $p\big(\bdelta^{(d)}\mid\mathbf W^{(d)},\tilde{\mathbf E}^{(d)},\Theta\big)$:
$q_\phi\big(\bdelta^{(d)}\mid\mathbf W^{(d)}\big)
=\Norm\big(\bdelta^{(d)};\bmu_\phi(\mathbf u^{(d)}),\operatorname{diag}(\bsigma_\phi^2(\mathbf u^{(d)}))\big)$,
where $\mathbf u^{(d)}\in\mathbb R^V$ is the $\ell_1$-normalized bag of words of the document and $\bmu_\phi,\bsigma_\phi$ are the outputs of a multilayer perceptron with parameters $\phi$, as in ETM~\citep{dieng2020topic}. The inference network conditions only on the words of the document; the contextual view enters the inference through the likelihood. The sentence topics require no variational approximation, since they are summed out exactly in~\Cref{eq:sentence_likelihood} and, once $\btheta^{(d)}$ is known, their posterior $\gamma_{mk}^{(d)}$ follows by normalizing its terms.

\textbf{Objective and optimization.}
For a document $d$, our objective is (derivation in Appendix~\ref{app:proof2}):
\begin{align}
\begin{aligned}
\mathcal L_d(\Theta,\phi)
&=\mathbb E_{q_\phi}\Bigg[\sum_{m=1}^{M^{(d)}}\log p\big(\mathbf w_m^{(d)},\tilde{\mathbf e}_m^{(d)}\mid\btheta^{(d)}\big)\Bigg]\\
&\phantom{=}-\KL\Big(q_\phi\big(\bdelta^{(d)}\mid\mathbf W^{(d)}\big)\,\Big\|\,\Norm(0,I_K)\Big),
\end{aligned}
\label{eq:elbo}
\end{align}
where $\mathbb E_{q_\phi}$ is the expectation under $q_\phi(\bdelta^{(d)}\mid\mathbf W^{(d)})$. The first term rewards topics that explain both views of every sentence; each summand is the logarithm of the sentence likelihood of~\Cref{eq:sentence_likelihood}, computed stably with the log-sum-exp operator. The second term keeps the variational distribution close to the prior and has the closed form $\frac12\sum_{k=1}^{K}\big(\sigma_{\phi,k}^2+\mu_{\phi,k}^2-1-\log\sigma_{\phi,k}^2\big)$.

We maximize $\sum_{d=1}^{D}\mathcal L_d$ by alternating two steps (\Cref{alg:inference}).
\textit{(i) Inference network and lexical parameters.} We update $\phi$ and $\Theta_{\text{lex}}$ by stochastic gradient ascent on mini-batches of documents, estimating the expectation in~\Cref{eq:elbo} with a single sample and the reparameterization $\bdelta^{(d)}=\bmu_\phi+\bsigma_\phi\odot\bm\epsilon$, with $\bm\epsilon\sim\Norm(0,I_K)$.
\textit{(ii) Contextual parameters.} At the end of each epoch, we compute the posteriors $\gamma_{mk}^{(d)}$ of all sentences by normalizing the terms of~\Cref{eq:sentence_likelihood}, using the mean of $q_\phi$, $\btheta^{(d)}=\sm\big(\bmu_\phi(\mathbf u^{(d)})\big)$, and update $\Theta_{\text{ctx}}$ in closed form. For fixed posteriors, the maximum-likelihood estimates of the parameters of~\Cref{eq:hddc_covariance} are: the estimate of $\mathbf V_k$ collects the $h_k$ eigenvectors of $S_k = \frac{1}{n_k}\sum_{d,m}\gamma_{mk}^{(d)}\big(\tilde{\mathbf e}_m^{(d)}-\hat\bmu_k\big)\big(\tilde{\mathbf e}_m^{(d)}-\hat\bmu_k\big)^\top$ and
\begin{align}
\hat \sigma_k^2=\frac{1}{E-h_k}\sum_{j=h_k+1}^{E}\hat\lambda_{kj}=\frac{\operatorname{tr}(\mathbf S_k)-\sum_{j=1}^{h_k}\hat\lambda_{kj}}{E-h_k},
\label{eq:hddc_sigma}
\end{align}
where $\hat\lambda_{k1}\ge\dots\ge\hat\lambda_{kE}$ are the ordered eigenvalues of $\mathbf S_k$, $\hat\bmu_k = \frac{1}{n_k}\sum_{d,m}\gamma_{mk}^{(d)}\,\tilde{\mathbf e}_m^{(d)}$ and $n_k=\sum_{d,m}\gamma_{mk}^{(d)}$.
We update the contextual parameters in closed form, conditional on the current responsibilities $\gamma_{mk}^{(d)}$. This update requires no learning rate and keeps $\hat{\mathbf V}_k$ orthonormal by construction. The contextual parameters are initialized by running k-means on all normalized sentence embeddings and applying the same updates with hard assignments, $\gamma_{mk}^{(d)}\in\{0,1\}$.
After training, we compute $\btheta^{(d)}$ from the mean of $q_\phi$ and assign each sentence to $\arg\max_k\gamma_{mk}^{(d)}$.
In practice, we use in all experiments the same subspace dimension for every topic, i.e., $h_k = h$ for all $k$.
\begin{figure*}[t]
\centering
\definecolor{topicA}{HTML}{0072B2}\colorlet{hlA}{topicA!30}
\definecolor{topicB}{HTML}{E69F00}\colorlet{hlB}{topicB!30}
\definecolor{topicC}{HTML}{009E73}\colorlet{hlC}{topicC!30}
\definecolor{topicD}{HTML}{CC79A7}\colorlet{hlD}{topicD!30}
\definecolor{topicE}{HTML}{56B4E9}\colorlet{hlE}{topicE!30}
\def\sentA#1{{\sethlcolor{hlA}\hl{#1}}}
\def\sentB#1{{\sethlcolor{hlB}\hl{#1}}}
\def\sentC#1{{\sethlcolor{hlC}\hl{#1}}}
\def\sentD#1{{\sethlcolor{hlD}\hl{#1}}}
\def\sentE#1{{\sethlcolor{hlE}\hl{#1}}}
\def\wrongsent#1#2{{\color{red!75!black}\itshape\sethlcolor{hl#1}\hl{#2}}}
\begin{minipage}{1\linewidth}
\scriptsize
\sentA{US consumer confidence up..Consumers' confidence in the state of the US economy is at its highest for five months and they are optimistic about 2005, an influential survey says...The feel-good factor among US consumers rose in December for the first time since July according to new data.} \sentA{US retailers have reported strong sales over the past 10 days after a slow start to the crucial festive season. According to figures also released on Tuesday, sales in shopping malls in the week to 25 December were 4.3\% higher than in 2003 following a last-minute rush.} \sentB{Career honor for actor DiCaprio. Actor Leonardo DiCaprio's ``exceptional career'' has been honored at the Santa Barbara International Film Festival. The star was presented with the award by Martin Scorsese, who directed him in Oscar-nominated movie The Aviator.} \wrongsent{D}{``It's what I want to do for the rest of my life.''} 
\sentC{[$\dots$] Andy Williams, chief executive of Powys Local Health Board, had said it was ``a totally unacceptable way to behave''...Mr Williams had said he did not think it was a Welsh-English issue, but said Hereford hospital was ``struggling financially and trying to pass the problem back to Powys''.} \sentD{Chelsea denied by James heroics..A brave defensive display, led by keeper David James, helped Manchester City hold the leaders Chelsea...After a quiet opening, James denied Damien Duff, Jiri Jarosik and Mateja Kezman, while Paul Bosvelt cleared William Gallas' header off the line.} \sentD{Chelsea had most of the possession in the second half but James kept out Frank Lampard's free-kick and superbly tipped the same player's volley wide.} \sentD{City went into the game with the proud record of being the only domestic team to beat Chelsea this season. [$\dots$]} 
\sentE{[$\dots$] The new technology will be welcomed by Hollywood film studios which are increasingly relying on revenue from DVD sales...The film industry has stepped up efforts to fight DVD piracy in the last 12 months, taking legal action against websites which offer pirated copies of DVD movies for download...``Ultimately, we see RipGuard DVD... evolving beyond anti-piracy, and towards enablement of legitimate online transactions, interoperability in tomorrow's digital home, and the upcoming high-definition formats,'' said Steve Weinstein, executive vice president and general manager of Macrovision's Entertainment Technologies Group.}
\end{minipage}
\caption{\textbf{SentenceETM sentence-level topic assignments}. BBC News, $K=5$. Background color marks the predicted topic (\sentA{business} / \sentB{entertainment} / \sentC{politics} /
\sentD{sport} / \sentE{tech}); misassignment in red.}
\label{fig:bbc_k_5}
\end{figure*}

\section{Experimental Setting}
\label{sec:exp_setting}
Prior work evaluates sentence-level topic models based on their induced document-level topic representations and therefore does not directly assess how topics are distributed across sentences. For example, senLDA~\citep{balikas2016topic} is evaluated using held-out perplexity and the $F_1$ score of a classifier trained on document topic proportions, while SentenceLDA~\citep{cha-lee-2024-sentencelda} is evaluated based on the quality of the document representations it induces. We argue that evaluating sentence-level topic models requires, first and foremost, datasets with sentence-level ground-truth topic labels. In the absence of suitable existing datasets, we propose a synthetic corpora generation strategy that provides such labels while allowing us to control the document-level topic mixtures. We remind readers that the labels are used only afterward for model evaluation and are not used during training (all methods are unsupervised).
Details on the used datasets, model training, and metrics in Appendix~\ref{app:exp_setting}.

\textbf{Synthetic corpora.} Let $L$ be the number of sentences of the desired document to create, $K$ the number of topics it mixes, and $p$ the percentage of the document that belongs to its dominant topic. Each synthetic document is assembled by drawing $m_1 = p\cdot L$ sentences from a document of the dominant topic and $m_2 = L(1-p) /(K-1)$ sentences from (random) documents of each of the remaining $K-1$ topics, all drawn from a corpus whose documents carry a single class label. The new document mixture is therefore \textit{fixed by construction}.
Varying $K$ at fixed $L$ and $p$ therefore varies how the same $L\cdot (1 - p)$ non-dominant-topic sentences are \textit{fragmented}: at $K=2$ they form a single block of $L\cdot (1 - p)$ sentences, at the largest admissible $K$ they are $L\cdot (1 - p)$ isolated sentences, one per topic. Three constraints delimit the admissible $K$. Every topic must contribute at least one sentence, $m_2 \geq 1$, hence $K \leq 1 + L(1-p)$; the dominant topic must dominate, $m_1 > m_2$, i.e., $p > 1/K$; and $K \leq K_{\max}$, the number of classes in the source corpus.

We build the synthetic corpora from: BBC News~\citep{greene2006practical}, 20 Newsgroups~\footnote{\url{http://qwone.com/~jason/20Newsgroups/}}, and Wikipedia~\citep{hoyle2025proxann,merity2016pointer}. 
We set $L = 20$ and $p = 0.6$ for all three corpora. The dominant topic then contributes $m_1 = p\cdot L = 12$ sentences, which must come from a single article, so $L$ is bounded by the length of the source articles: $m_1 = 12$ is not binding for Wikipedia (median $\sim110$ sentences per article), is available in about half of the BBC articles (median $\sim13$), and in about a quarter of the 20 Newsgroups posts, whose lengths are heavily skewed (median $\sim6$, mean $11.1$). A larger $L$ would leave too few eligible articles in BBC and 20 Newsgroups, while a smaller one would make sentence topics harder to recover, even at $K=2$. With $p = 0.6$, the dominant topic covers a clear majority of every document, while the remaining $L\cdot(1-p) = 8$ sentences allow the non-dominant mass to be split across up to eight topics.
We use $K \in \{2,3,5,9\}$ for 20 Newsgroups and Wikipedia, and $K \in \{2,3,5\}$ for BBC News (as it has a maximum of $5$ classes). The synthetic corpus has $700$ documents, each of $20$ sentences.

\textbf{Political debates dataset.}
ElecDeb60to20~\citep{goffredo2023argument} comprises 44 U.S. televised debates (1960--2020), collected from the Commission on Presidential Debates website\footnote{\url{http://www.debates.org/}}, which provides transcripts of presidential and vice-presidential debates. For our experiments, we consider the debate between Biden and Trump on the 29th of October 2020, as it is the longest. Specifically, the debates are organized into 6 segments (see Appendix~\ref{app:exp_setting}), each presenting a main topic, along with the journalists' and the candidates' answers. 

\textbf{Baselines.} We test {\sentetm} against the main candidates of each of the three families of topic models. At the word-level, we consider LDA~\citep{blei2003latent} and ETM~\citep{dieng2020topic}; at the document-level, BERTopic~\citep{grootendorst2022bertopic}, which clusters document embeddings, and finally at the sentence-level, senLDA~\citep{balikas2016topic}, an LDA in which all words of a sentence share one topic, and SentenceLDA~\citep{cha-lee-2024-sentencelda}, a Gaussian LDA over sentence embeddings (cf.~\Cref{sec:related}). All models use the same preprocessing and vocabulary, and all embedding-based models (BERTopic, SentenceLDA, {\sentetm}) use the same sentence encoder, either \texttt{all-MiniLM-L6-v2} or \texttt{all-mpnet-base-v2}.
Our evaluation requires one topic per sentence, which word- and document-level models do not output directly. Therefore, we derive sentence-level assignments as follows: for LDA and ETM, every word $w$ of document $d$ takes its most probable topic given the fitted model, $\arg\max_k \theta_{dk}\beta_{kw}$, and a sentence takes the topic receiving the majority of its word assignments. For BERTopic, each document is represented by the mean embedding of its sentences, using the same encoder as the sentence-level models to avoid truncation and give each sentence equal weight. Documents are clustered with UMAP and HDBSCAN; the resulting clusters are reduced to $K$ topics, outliers are reassigned to the nearest topic, and every sentence inherits the topic of its document.

\begin{figure}[!t]
    \centering
    \includegraphics[width=\columnwidth]{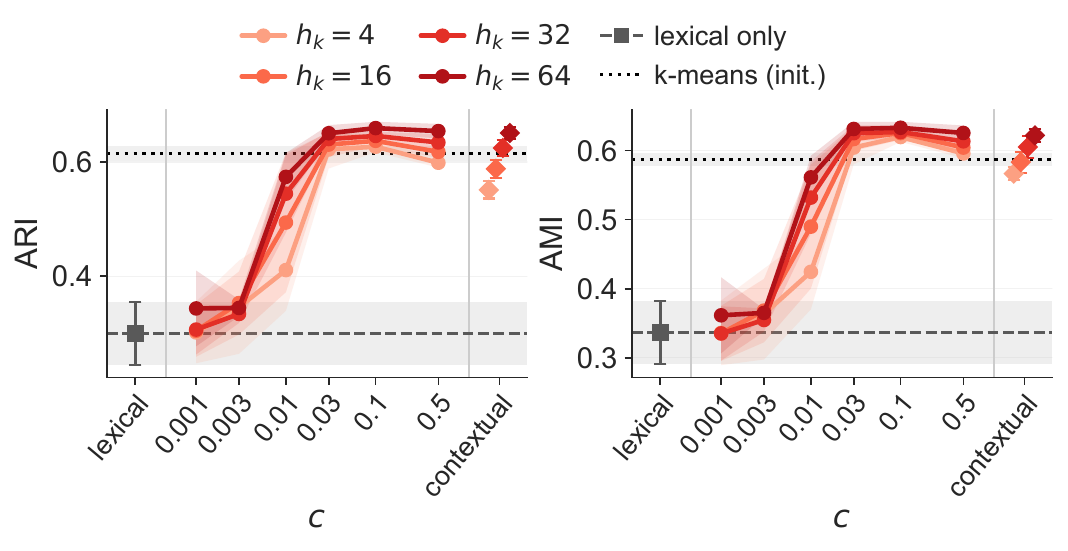}
\caption{\textbf{Ablation on BBC News} ($K=5$, \texttt{all-mpnet-base-v2}).
ARI and AMI as a function of $c$, for the common subspace
dimensions $h_k=h\in\{4,16,32,64\}$.
Lexical evidence only in gray; diamonds denote contextual-only
models.}
    \label{fig:ablation}
\end{figure}
\begin{figure}[!t]
\centering
\includegraphics[width=\columnwidth]{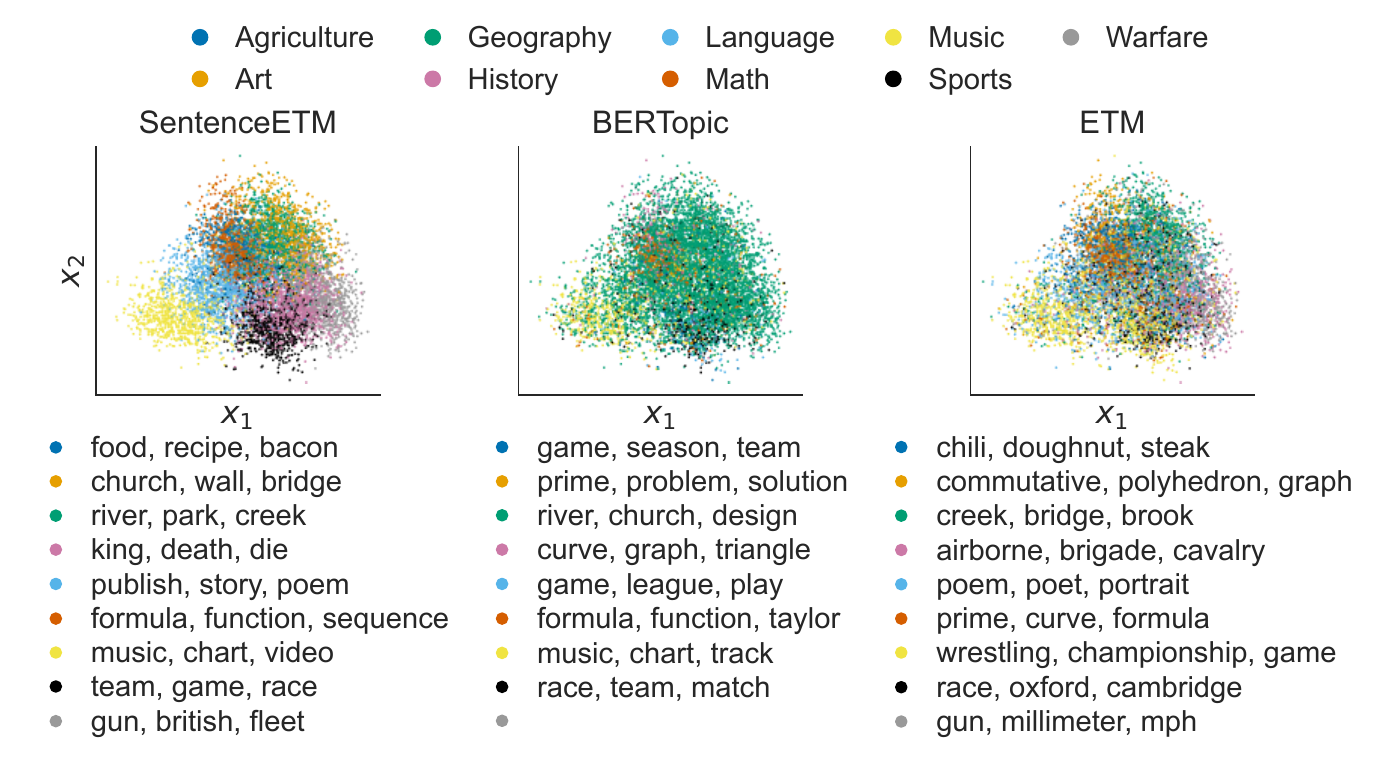}

\caption{\textbf{Sentence-level topic assignments} (Wikipedia, \texttt{all-mpnet-base-v2}, $K=9$). Sentences projected onto the first two embedding PCs and colored by assigned topic; top-3 words below each panel.}
\label{fig:embeddings_wiki}
\end{figure}

\section{Experimental Results}
\label{sec:exp_results}
\Cref{fig:bbc_k_5} shows a randomly selected document for $K=5$ on BBC News synthetic (cf.~\Cref{sec:exp_setting}). {\sentetm} switches topics from one sentence to the next according to the actual content. The red sentence ``\textit{It's what I want to do for the rest of my life}'' highlights the model mistake. Notably, such a sentence carries no topic-specific information and is therefore not representative of any of the topics. Moreover, in~\Cref{fig:embeddings_wiki} we show the structure of the latent space for a randomly chosen run on Wikipedia (full details in Appendix~\ref{app:latent}). Finally, we refer the reader to Appendix~\ref{app:original} for results on the original datasets.
\subsection{Results on synthetic corpora}
Additional analysis of the model, including sensitivity analyses for the sentence encoder and document length, is available in Appendices~\ref{app:encoder} to~\ref{app:doc_length}. Computational cost in Appendix~\ref{app:cost}. 
SenClu~\citep{schneider2023efficient} is in~\Cref{app:senclu} due to space constraints.

\textbf{Ablation study.}
To assess the contribution of the lexical and contextual evidence, we vary the weight $c$ of the contextual likelihood and a common subspace
dimension $h$, setting $h_k=h$ for all topics. We consider two reference models: one that uses lexical evidence only ($c=0$, where the subspace dimension plays no role) and one that uses contextual evidence only. Since {\sentetm} initializes its contextual component from a $k$-means partition of the normalized sentence embeddings, we additionally report this partition as a strong contextual baseline. We use BBC News with $K=5$ and \texttt{all-mpnet-base-v2}, and repeat each configuration on 3 synthetic corpora sampled with different seeds.~\Cref{fig:ablation} reports ARI and AMI as a function of $c$ for each $h_k$. For $c\le 0.003$, the contextual evidence has almost no influence: the model behaves like the lexical-only reference (ARI $\approx 0.30$) and varies strongly across corpora. Between $c=0.003$ and $c=0.03$, the contextual evidence takes over, earlier for larger $h_k$, whose richer subspaces make it more informative. For $c\ge 0.03$, performance is largely stable (ARI approximately $0.62$--$0.66$), with the best results around $c=0.1$ for every tested subspace dimension. The gain over lexical evidence alone is large (about $+0.33$ ARI), whereas the gain over contextual evidence alone depends on $h_k$: it is about $+0.08$ for $h_k=4$ and progressively decreases as the subspace becomes more expressive, down to about $+0.01$ for $h_k=64$. Thus, lexical and contextual evidence are complementary: lexical information contributes most when the contextual model is constrained, while for larger subspaces the contextual representation already explains much of the sentence structure. The lexical component additionally yields the topic--word distributions used to interpret the topics. The setting used in the main experiments ($h_k=32$ for all $k$, $c=0.5$) lies within the stable region. The AMI results show the same behavior.

\begin{figure}[!tb]
\centering
\begin{subfigure}{\columnwidth}
    \centering
    \includegraphics[width=\columnwidth]{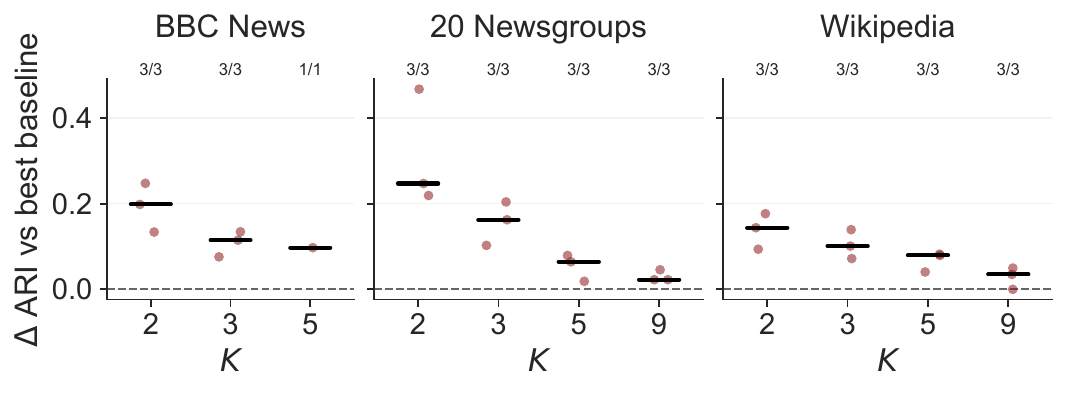}
    \caption{ARI}
    \label{fig:ari_gain}
    \end{subfigure}
\hfill
\begin{subfigure}{\columnwidth}
    \centering
    \includegraphics[width=\columnwidth]{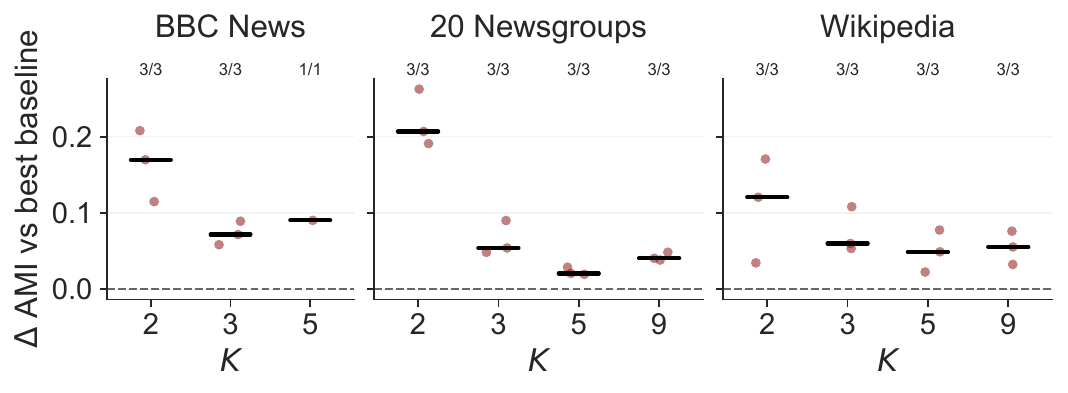}
    \caption{AMI}
    \label{fig:ami_gain}
\end{subfigure}
    \caption{\textbf{{\sentetm} per-subset gain, \texttt{all-mpnet-base-v2}.} Each point corresponds to a subset of (K) ground-truth classes, with ($\Delta$) measured relative to the best-performing baseline for that subset (SentenceLDA) and averaged over seeds.} 
    \label{fig:gain}
\end{figure}

\begin{figure}[htbp!]
\begin{subfigure}{\columnwidth}
    \centering
    \includegraphics[width=\columnwidth]{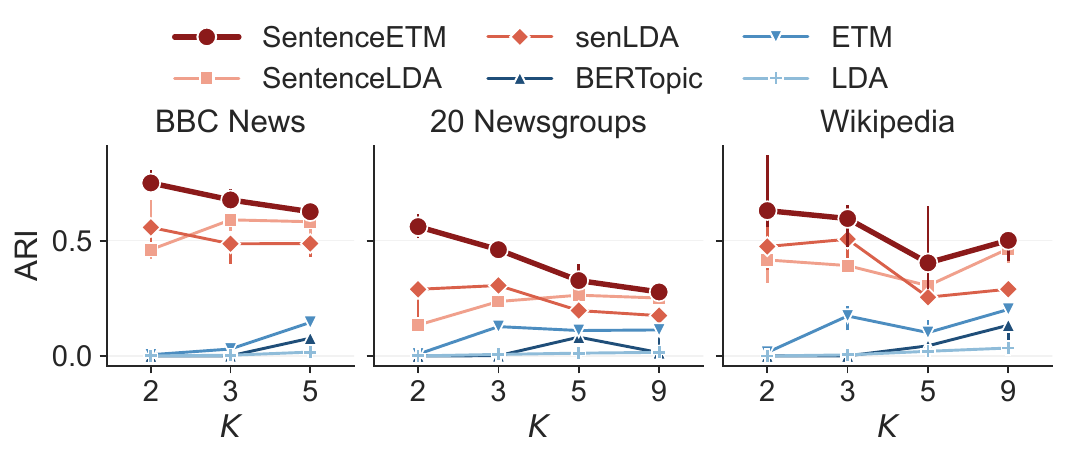}
    \caption{ARI}
\label{fig:ari}
\end{subfigure}
\hfill
\begin{subfigure}{\columnwidth}
    \centering
    \includegraphics[width=\columnwidth]{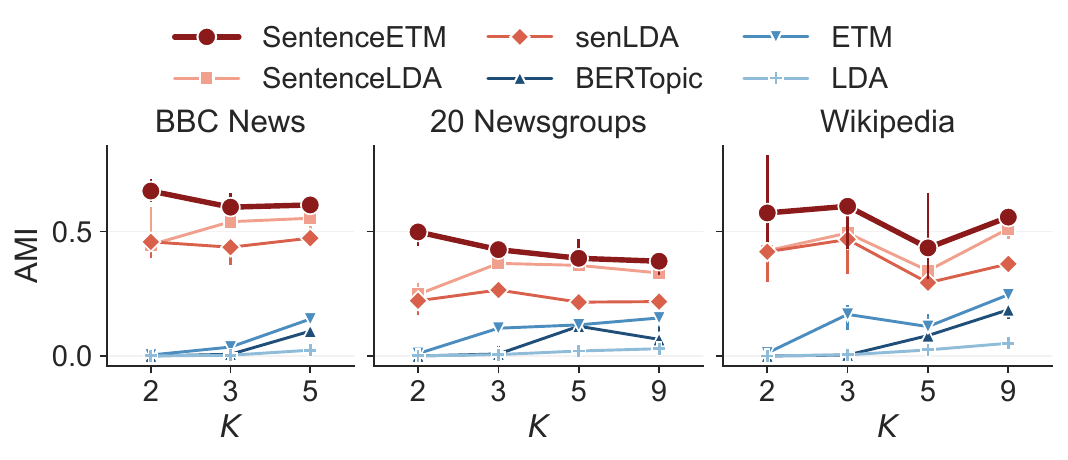}
    \caption{AMI}
\label{fig:ami}
\end{subfigure}
    \caption{\textbf{ARI and AMI as function of the number of topics $K$, \texttt{all-mpnet-base-v2}}. Markers indicate the median across runs, while bars show the interquartile range. Higher is better.}
    \label{fig:main_metric}
\end{figure}

\textbf{Analysis in terms of ARI and AMI.}~\Cref{fig:main_metric} reports clustering quality analysis as a function of the number of topics $K$ (\texttt{all-mpnet-base-v2}). The complete set of numerical results is available in~\Cref{tab:results_bbc,tab:results_20k,tab:results_wiki}. Two groups of methods clearly emerge. {\sentetm}, senLDA, and SentenceLDA, the other baselines lag behind, as their ARI and AMI are close to zero across all three datasets and increase only slightly as $K$ grows. This follows from the construction of the corpora: at $K=2$, every document contains both topics in proportions $0.6/0.4$, so document-level information provides little information about individual sentences; as $K$ grows, the dominant topic stands out more clearly against each of the others, making document-level assignments marginally informative. Overall, the proposed model achieves the best ARI and AMI across almost all settings, regardless of the encoder or dataset. A substantial part of the variability in the results comes from the choice of class subset, indeed some combinations of classes are intrinsically easier to separate than others (see~\Cref{tab:bbc_detail,tab:20k_detail,tab:wiki_detail}). To isolate the contribution of the model,~\Cref{fig:gain} compares methods on the same subsets. For each subset of $K$ classes, we average scores over seeds and compute $\Delta = s_{\text{{\sentetm}}} - \max_{b} s_b$, where the maximum is taken over all baselines on that subset. A positive $\Delta$ means that {\sentetm} outperforms every baseline on that subset. In both ARI and AMI, {\sentetm} outperforms all baselines across all 31 subsets, with median gains ranging from about 0.03 to 0.2, depending on the dataset and $K$. 
The gain is largest at small $K$ and decreases as $K$ grows, as the strongest baseline, typically SentenceLDA, becomes more competitive.
The most challenging dataset turned out to be 20 Newsgroups, possibly due to its less-curated, more conversational nature. Unlike BBC News and Wikipedia, whose documents are generally centered on a well-defined subject, 20 Newsgroups consists of user-written discussion posts that may include digressions, quoted text, or generic conversational content. As a result, individual sentences can be only weakly indicative of the document-level topic.

\subsection{Results on ElecDeb60to20}
We illustrate {\sentetm} 's flexibility by applying it to a specific use case. We consider the first 2020 U.S. presidential debate between Donald Trump and Joe Biden, in which the moderator divided the discussion into six segments, and treat each speaking turn as a document and its sentences as the units to which topics are assigned.
Because {\sentetm} treats the sentence encoder as a separate, interchangeable component, the encoder can be chosen to suit the application. Here, we want topics to capture \textit{what} is discussed rather than \textit{how} or \textit{by whom}: a sentence attacking an opponent's record on the economy and one defending one's own should fall under the same topic. We therefore use the instruction-following encoder \texttt{Qwen3-Embedding-0.6B} and prepend to every sentence the instruction ``\textit{Given a sentence spoken in a U.S. presidential debate, identify the policy issue it discusses, regardless of the speaker's stance or tone}''. We consider $K=20$ topics and an extended stopword list that includes words specific to spoken debate.
\begin{figure}[!t]
    \centering
    \includegraphics[width=\columnwidth]{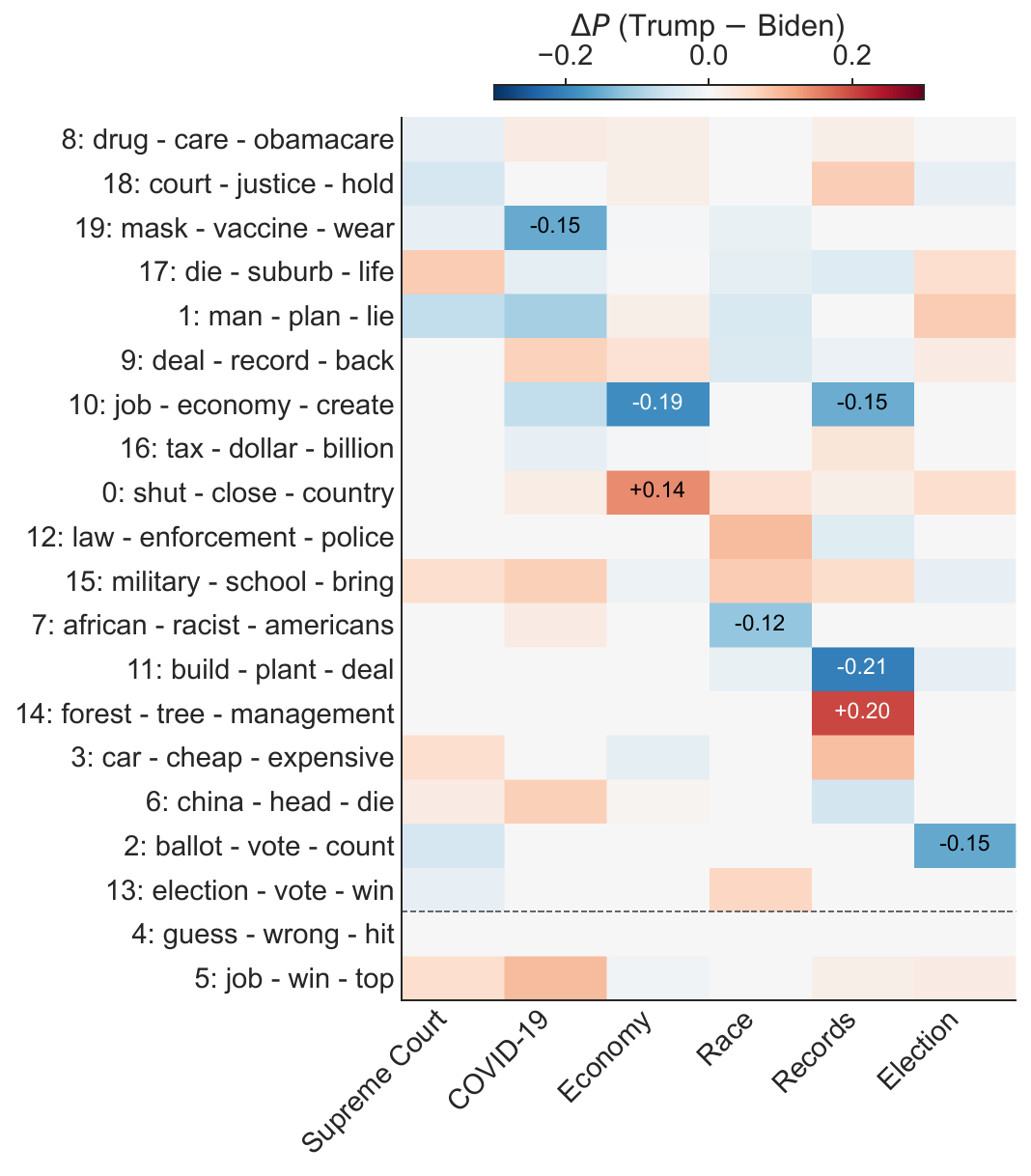}
\caption{
\textbf{Difference in probabilities of topic use between Trump and Biden across debate segments.} $\Delta P = P(T \mid S, \text{Trump}) - P(T \mid S, \text{Biden})$.
}
    \label{fig:debate_segments}
\end{figure}

\textbf{Topics across debate segments.}
\Cref{fig:debate_segments} compares how the two candidates distribute their sentences over topics within each segment. For each segment and topic, we compute the share of each candidate's sentences assigned to that topic and report the difference $\Delta P$ (Trump $-$ Biden); red cells indicate topics used more by Trump, blue cells indicate topics used more by Biden, and differences of at least 0.10 are annotated. The two topics below the dashed line capture generic conversational sentences. The largest differences reveal different emphases within the same segment. On race, 
Biden speaks more about racism towards African Americans ($-0.12$). On the economy, Biden focuses more on jobs and businesses ($-0.19$), while Trump discusses 
lockdowns (\textit{shut, country, close}, $+0.14$). On election integrity, Biden is on voting and ballot counting ($-0.15$). 
In the COVID-19 segment, Biden devotes sentences to masks and vaccines ($-0.13$), while Trump emphasizes the timing of the response and comparisons with other countries (top-words in Appendix~\ref{app:top_words_debate}).   

\textbf{Topic diversity within speaking turns.}
\begin{figure}[!t]
    \centering
    \includegraphics[width=\columnwidth]{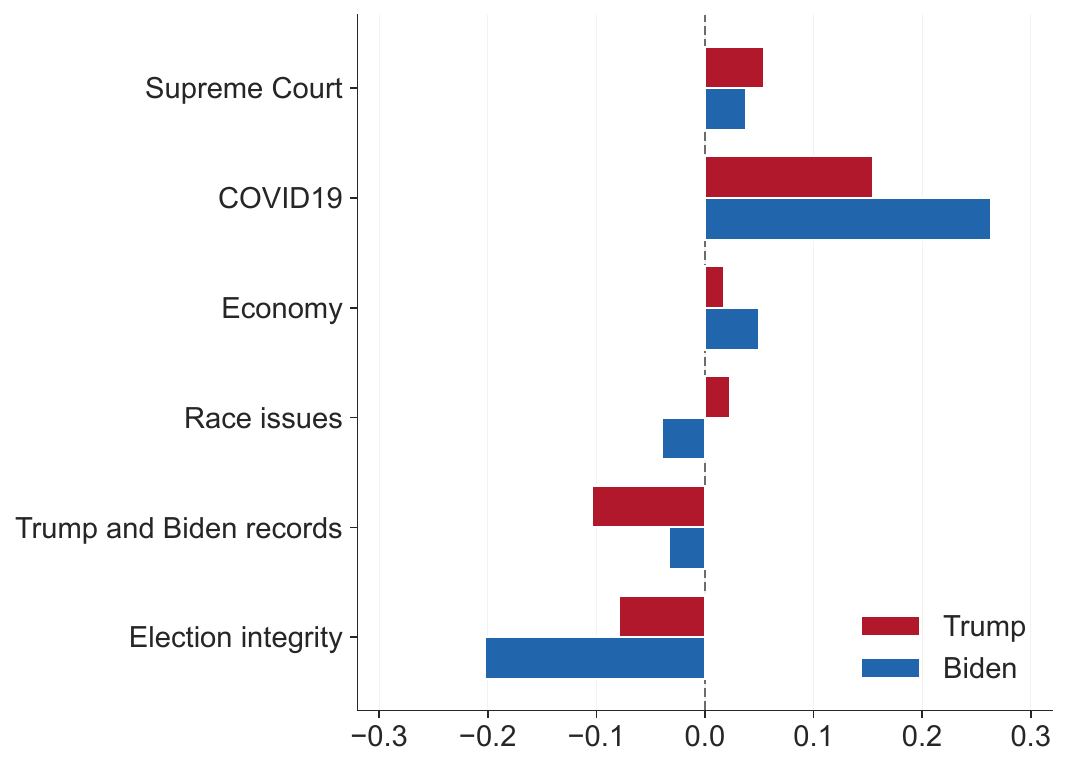}
\caption{
\textbf{New topics per additional sentence within speaking turns}, by segment and candidate, as a difference from each candidate's average. Positive values: more topics than usual; negative values: fewer.
}
    \label{fig:entropy_debate}
\end{figure}
\Cref{fig:entropy_debate} measures how often a candidate introduces new topics within a speaking turn. Within a turn, we say that a sentence introduces a new topic if its topic has not appeared in any previous sentence of that turn. The first sentence of a turn is excluded, since it cannot introduce a new topic. The index is the fraction of the remaining sentences that introduce a new topic, computed over all the turns of a candidate in a segment:
$\frac{\sum_{d}(T^{(d)}-1)}{\sum_{d}(M^{(d)}-1)}$,
where $M^{(d)}$ is the number of sentences of turn $d$ and $T^{(d)}$ the number of distinct topics it contains. The index is 0 when every turn stays on a single topic and 1 when every sentence introduces a new topic.
Computed over all the turns of the debate, the index is almost identical for the two candidates (0.45 for Trump, 0.43 for Biden): around two sentences out of five introduce a new topic. We use this value as each candidate's average, and the figure reports, for each segment, the difference between the candidate's index in that segment and their average. A positive value means that, in that segment, the candidate introduces new topics more often than usual; a negative value indicates lower within-turn topic diversity. Differences are larger across segments than between speakers. COVID-19 is the least cohesive segment for both candidates ($+0.15$ for Trump, $+0.26$ for Biden). Election integrity is the most cohesive segment for Biden ($-0.20$), whose turns stay focused on voting, while Trump's most cohesive segment is the candidates' records ($-0.10$). 

\section{Conclusion}
We introduced {\sentetm}, a sentence-level topic model that combines lexical evidence from topic--word distributions with contextual evidence from topic-specific low-dimensional subspaces of the sentence-embedding space. Document-level topic proportions act as a prior, and Gaussian parameters are updated in closed form. We also proposed a synthetic corpus protocol in which each sentence has a known topic.
{\sentetm} achieves the best ARI and AMI in almost all settings and remains robust to a smaller sentence encoder, unlike SentenceLDA. In a U.S. presidential debate, it reveals differences in topic emphasis between candidates and within-turn topic diversity. 
Future work includes modeling sentence order and topic transitions, and complementing synthetic labels with human annotations.

\subsection*{AI use statement}
In this work, we used generative AI tools for tasks requiring disclosure, including: \emph{method implementation}, by drafting portions of the code, assisting with notebooks, and helping integrate existing state-of-the-art implementations; \emph{interpretation of results}, by discussing experimental outcomes and possible explanations reported in the paper; \emph{qualitative analysis}, by proposing short labels for the debate topics in \Cref{tab:debate_top_words}.
\textbf{The model itself, including the use of high-dimensional Gaussian models for contextual evidence, was conceived by the authors. Generative AI was not used to generate synthetic data, formulate hypotheses or novel mathematical claims, or clean the datasets}.
All AI-assisted material was reviewed by the authors. All code was inspected and executed by the authors; all numerical results reported in the paper were obtained from experimental outputs rather than generated by AI.
We take full responsibility for the final content of this work, including any text, claims, code, or other artifacts produced with the assistance of generative AI.





\bibliographystyle{apalike}
\bibliography{aistats_conference}

\clearpage
\appendix
\thispagestyle{empty}

\onecolumn
\aistatstitle{Supplementary Materials}
\section{Proofs}
This section provides the proofs supporting~\Cref{sec:method}. We also provide algorithms summarizing both the generative process and the inference procedure.
\subsection{ELBO decomposition}
\label{app:proof1}
By the product rule,
\begin{align*}
p\big(\mathbf W^{(d)},\tilde{\mathbf E}^{(d)},\bdelta^{(d)}\mid\Theta\big)
= p\big(\mathbf W^{(d)},\tilde{\mathbf E}^{(d)}\mid\Theta\big)\,
  p\big(\bdelta^{(d)}\mid\mathbf W^{(d)},\tilde{\mathbf E}^{(d)},\Theta\big).
\end{align*}
Therefore, for every value of $\bdelta^{(d)}$ at which the posterior is positive,
\begin{align*}
\log p\big(\mathbf W^{(d)},\tilde{\mathbf E}^{(d)}\mid\Theta\big)
= \log p\big(\mathbf W^{(d)},\tilde{\mathbf E}^{(d)},\bdelta^{(d)}\mid\Theta\big)
- \log p\big(\bdelta^{(d)}\mid\mathbf W^{(d)},\tilde{\mathbf E}^{(d)},\Theta\big).
\end{align*}
Taking the expectation of both sides under $q$ leaves the left-hand side unchanged, as it does
not depend on $\bdelta^{(d)}$:
\begin{align*}
\log p\big(\mathbf W^{(d)},\tilde{\mathbf E}^{(d)}\mid\Theta\big)
&= \mathbb E_q\Big[\log p\big(\mathbf W^{(d)},\tilde{\mathbf E}^{(d)},\bdelta^{(d)}\mid\Theta\big)\Big]\\
&\quad- \mathbb E_q\Big[\log p\big(\bdelta^{(d)}\mid\mathbf W^{(d)},\tilde{\mathbf E}^{(d)},\Theta\big)\Big].
\end{align*}
Finally, adding and subtracting $\mathbb E_q[\log q(\bdelta^{(d)})]$ and regrouping,
\begin{align*}
\log p\big(\mathbf W^{(d)},\tilde{\mathbf E}^{(d)}\mid\Theta\big)
&= \mathbb E_q\bigg[\log\frac{p\big(\mathbf W^{(d)},\tilde{\mathbf E}^{(d)},\bdelta^{(d)}\mid\Theta\big)}{q\big(\bdelta^{(d)}\big)}\bigg]
 + \mathbb E_q\bigg[\log\frac{q\big(\bdelta^{(d)}\big)}{p\big(\bdelta^{(d)}\mid\mathbf W^{(d)},\tilde{\mathbf E}^{(d)},\Theta\big)}\bigg]\\
&= \mathcal L\big(q;\Theta\big)
 + \KL\Big(q\big(\bdelta^{(d)}\big)\,\Big\|\,p\big(\bdelta^{(d)}\mid\mathbf W^{(d)},\tilde{\mathbf E}^{(d)},\Theta\big)\Big).
\end{align*}

\subsection{Derivation of~\Cref{eq:elbo}}
\label{app:proof2}
Recall from the ELBO decomposition that
\begin{equation*}
\mathcal L(q;\Theta)=\mathbb E_q\Big[\log p\big(\mathbf W^{(d)},\tilde{\mathbf E}^{(d)},\bdelta^{(d)}\mid\Theta\big)-\log q\big(\bdelta^{(d)}\big)\Big].
\end{equation*}
Moreover,
\begin{equation*}
\log p\big(\mathbf W^{(d)},\tilde{\mathbf E}^{(d)},\bdelta^{(d)}\mid\Theta\big)
=\log\Norm\big(\bdelta^{(d)};0,I_K\big)+\sum_{m=1}^{M^{(d)}}\log p\big(\mathbf w_m^{(d)},\tilde{\mathbf e}_m^{(d)}\mid\btheta^{(d)}\big).
\end{equation*}
Substituting into $\mathcal L(q;\Theta)$, using the linearity of the expectation and the
definition of the Kullback--Leibler divergence, $\KL(q\,\|\,p)=\mathbb E_q[\log q-\log p]$,
\begin{align*}
\mathcal L(q;\Theta)
&=\mathbb E_q\bigg[\log\Norm\big(\bdelta^{(d)};0,I_K\big)+\sum_{m=1}^{M^{(d)}}\log p\big(\mathbf w_m^{(d)},\tilde{\mathbf e}_m^{(d)}\mid\btheta^{(d)}\big)-\log q\big(\bdelta^{(d)}\big)\bigg]\\
&=\mathbb E_q\bigg[\sum_{m=1}^{M^{(d)}}\log p\big(\mathbf w_m^{(d)},\tilde{\mathbf e}_m^{(d)}\mid\btheta^{(d)}\big)\bigg]+\mathbb E_q\Big[\log\Norm\big(\bdelta^{(d)};0,I_K\big)-\log q\big(\bdelta^{(d)}\big)\Big]\\
&=\mathbb E_q\bigg[\sum_{m=1}^{M^{(d)}}\log p\big(\mathbf w_m^{(d)},\tilde{\mathbf e}_m^{(d)}\mid\btheta^{(d)}\big)\bigg]-\KL\Big(q\big(\bdelta^{(d)}\big)\,\Big\|\,\Norm\big(0,I_K\big)\Big),
\end{align*}
which, with $q=q_\phi$, is~\Cref{eq:elbo}.

\subsection{Computing the contextual likelihood}
\label{app:hddc}
The covariance of~\Cref{eq:hddc_covariance} can be equivalently written as
\begin{align*}
\bSigma_k=\mathbf V_k\operatorname{diag}(\lambda_{k1},\dots,\lambda_{kh_k})\mathbf V_k^\top +\sigma_k^2\big(\mathbf I_E-\mathbf V_k\mathbf V_k^\top\big),
\end{align*}
since $\mathbf V_k\operatorname{diag}(\sigma_k^2,\dots,\sigma_k^2)\mathbf V_k^\top=\sigma_k^2\mathbf V_k\mathbf V_k^\top$. As $\mathbf V_k$ has orthonormal columns, $\bSigma_k$ has eigenvalues $\lambda_{k1},\dots,\lambda_{kh_k}$ along the columns of $\mathbf V_k$ and $\sigma_k^2$ along the $E-h_k$ orthogonal directions. Hence
\begin{align*}
\bSigma_k^{-1}&=\mathbf V_k\operatorname{diag}\big(\lambda_{k1}^{-1},\dots,\lambda_{kh_k}^{-1}\big)\mathbf V_k^\top
+\sigma_k^{-2}\big(\mathbf I_E-\mathbf V_k\mathbf V_k^\top\big),\\
\log|\bSigma_k|&=\sum_{j=1}^{h_k}\log\lambda_{kj}+(E-h_k)\log\sigma_k^2 .
\end{align*}
Let $\mathbf r=\tilde{\mathbf e}_m^{(d)}-\bmu_k$ and $\mathbf a=\mathbf V_k^\top\mathbf r\in\mathbb R^h_k$ be the coordinates of $\mathbf r$ in the topic subspace. Since $\|\mathbf r\|^2=\|\mathbf a\|^2+\|(\mathbf I_E-\mathbf V_k\mathbf V_k^\top)\mathbf r\|^2$,
\begin{align*}
\log\Norm\big(\tilde{\mathbf e}_m^{(d)};\bmu_k,\bSigma_k\big)
=-\frac12\Big[E\log 2\pi+\log|\bSigma_k|
+\sum_{j=1}^{h_k}\frac{a_j^2}{\lambda_{kj}}
+\frac{\|\mathbf r\|^2-\|\mathbf a\|^2}{\sigma_k^2}\Big].
\end{align*}
The third term measures the distance of the sentence from the topic mean within the topic subspace, weighted by the topic's variance along its directions; the fourth term measures its distance from the subspace, weighted by the noise variance $\sigma_k^2$.
The cost is $O(Eh_k)$ per sentence and topic, and no $E\times E$ matrix is ever formed. In practice, we lower-bound $\lambda_{kj}$ and $\sigma_k^2$ by a small constant to prevent degenerate topics.

\subsection{Generative process}
\label{app:generative}
\begin{algorithm}[h]
\caption{Generative process of {\sentetm}}
\label{alg:generative_process}
\begin{algorithmic}[1]
\For{each topic $k=1,\dots,K$}
  \State $\bm\beta_k\gets\sm(\brho^\top\balpha_k)$
  \State $\bSigma_k\gets\mathbf V_k\operatorname{diag}(\lambda_{k1}-\sigma_k^2,\dots,\lambda_{kh_k}-\sigma_k^2)\mathbf V_k^\top+\sigma_k^2\mathbf I_E$
\EndFor
\For{each document $d=1,\dots,D$}
  \State Draw $\bdelta^{(d)}\sim\Norm(0,I_K)$ and set $\btheta^{(d)}\gets\sm(\bdelta^{(d)})$
  \For{each sentence $m=1,\dots,M^{(d)}$}
    \State Draw the topic $z_m^{(d)}\sim\Mult(1;\btheta^{(d)})$
    \State Draw the embedding $\tilde{\mathbf e}_m^{(d)}\sim\Norm\big(\bmu_{z_m^{(d)}},\bSigma_{z_m^{(d)}}\big)$
    \For{each word $n=1,\dots,N_m^{(d)}$}
      \State Draw the word $w_{mn}^{(d)}\sim\Mult\big(1;\bm\beta_{z_m^{(d)}}\big)$
    \EndFor
  \EndFor
\EndFor
\end{algorithmic}
\end{algorithm}
\newpage
\subsection{Inference process}
\label{app:inference_process}
\begin{algorithm}[h]
\caption{Inference in {\sentetm}}
\label{alg:inference}
\begin{algorithmic}[1]
\Require documents $\{\mathbf W^{(d)}\}_{d=1}^{D}$ with sentence embeddings
$\{\mathbf e_m^{(d)}\}$; number of topics $K$;
topic-specific subspace dimensions $\{h_k\}_{k=1}^K$; number of epochs $T$

\State $\tilde{\mathbf e}_m^{(d)}
\gets\mathbf e_m^{(d)}/\|\mathbf e_m^{(d)}\|$ for every sentence

\State Run k-means with $K$ clusters on
$\{\tilde{\mathbf e}_m^{(d)}\}$;
set $\gamma_{mk}^{(d)}=1$ if sentence $m$ of $d$ is in cluster $k$, $0$ otherwise

\State Initialize $\Theta_{\mathrm{ctx}}$ from the hard assignments
using the closed-form estimates in~\Cref{sub_sec:inference};
initialize $\phi$ and $\Theta_{\mathrm{lex}}$ at random

\For{$t=1,\dots,T$}
  \For{each mini-batch $\mathcal B$ of documents}
    \For{$d\in\mathcal B$}
      \State $\bdelta^{(d)}
      \gets\bmu_\phi(\mathbf u^{(d)})
      +\bsigma_\phi(\mathbf u^{(d)})\odot\bm\epsilon$,
      \ $\bm\epsilon\sim\Norm(0,I_K)$
      \State $\btheta^{(d)}\gets\sm(\bdelta^{(d)})$
      \State Compute $\mathcal L_d$ with~\Cref{eq:elbo},
      summing out each $z_m^{(d)}$ in~\Cref{eq:sentence_likelihood}
    \EndFor

    \State Update $\phi$ and $\Theta_{\mathrm{lex}}$
    by a gradient step on $\sum_{d\in\mathcal B}\mathcal L_d$
    \Comment{$\Theta_{\mathrm{ctx}}$ fixed}
  \EndFor

  \State $\btheta^{(d)}
  \gets\sm\big(\bmu_\phi(\mathbf u^{(d)})\big)$ for every document
  \State Compute $\gamma_{mk}^{(d)}$ for every sentence
  \State Update $\Theta_{\mathrm{ctx}}$
  using the closed-form estimates in~\Cref{sub_sec:inference}
  \Comment{closed form}
\EndFor

\State \Return $\phi$, $\Theta$,
and the sentence topics $\arg\max_k\gamma_{mk}^{(d)}$
\end{algorithmic}
\end{algorithm}
\section{Additional Materials to Section~\ref{sec:exp_setting}}
\label{app:exp_setting}
\subsection{Synthetic dataset construction pipeline}
\begin{figure}[!htbp]
    \centering
    \includegraphics[width=.6\columnwidth]{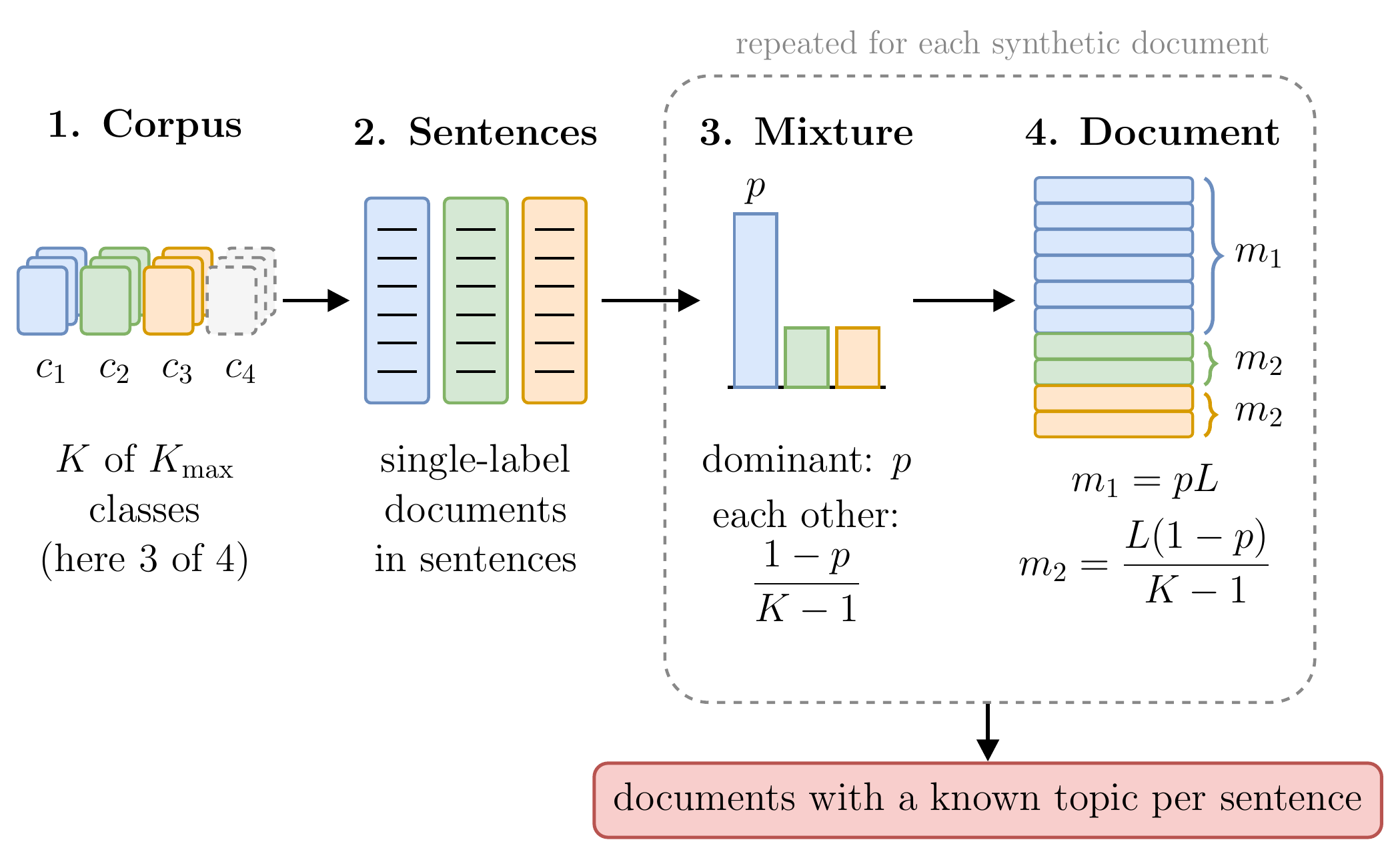}
\caption{\textbf{Synthetic corpora generation.} Example with $K=3$, $L=10$ and $p=0.6$.}
    \label{fig:synthetic_pipeline}
\end{figure}

The construction of a synthetic corpus proceeds in four steps, repeated independently for every run.
\begin{enumerate}
    \item As detailed in~\Cref{sec:exp_setting}, we start with existing topic modeling datasets that provide a single topic label per document: BBC News~\citep{greene2006practical}, 20 Newsgroups\footnote{\url{http://qwone.com/~jason/20Newsgroups/}}, and Wikipedia~\citep{merity2016pointer}. Specifically, 
    \begin{enumerate}
        \item BBC News dataset is a corpus consisting of 2.225 articles published on the BBC News website corresponding to 2004-2005 (in English). Each article is labeled under one of 5 categories: \texttt{business}, \texttt{entertainment}, \texttt{politics}, \texttt{sport} or \texttt{tech}.
        \item 20 Newsgroups dataset comprises around 18.000 newsgroups posts on 20 topics (in English):
        \texttt{rec.sport.hockey},
        \texttt{comp.sys.ibm.pc.hardware},
        \texttt{talk.politics.mideast},
        \texttt{comp.sys.mac.hardware},
        \texttt{sci.electronics},
        \texttt{talk.religion.misc},
        \texttt{sci.crypt},
        \texttt{sci.med},
        \texttt{alt.atheism},
        \texttt{rec.motorcycles},
        \texttt{rec.autos},
        \texttt{comp.windows.x},
        \texttt{comp.graphics},
        \texttt{sci.space},
        \texttt{talk.politics.guns},
        \texttt{misc.forsale},
        \texttt{rec.sport.baseball},
        \texttt{talk.politics.misc},
        \texttt{comp.os.ms-windows.misc},
        \texttt{soc.religion.christian}.
        \item We use the training split of the Wikipedia corpus from~\citet{hoyle2025proxann}, derived from WikiText-103~\citep{merity2016pointer}. 
        This dataset is a general-audience corpus consisting of $14{,}000$ English Wikipedia articles. We use the articles’ supercategory labels (15): \texttt{Language and literature}, \texttt{Natural sciences}, \texttt{Art and architecture}, \texttt{Music}, \texttt{Engineering and technology}, \texttt{Media and drama}, \texttt{History}, \texttt{Warfare}, \texttt{Sports and recreation}, \texttt{Geography and places}, \texttt{Social sciences and society}, \texttt{Philosophy and religion}, \texttt{Video games}, \texttt{Agriculture, food, and drink}, \texttt{Mathematics}.
        \end{enumerate}
    \item For a given $K$, we draw $K$ distinct topic labels from those available; each of the three runs uses a different subset. How $K$ is chosen is discussed in ~\Cref{sec:exp_setting}.

    \item For each of the $n$ documents, we draw the dominant topic uniformly among the $K$ selected labels. We then sample one source article of that topic among those with at least $m_1$ sentences, and take $m_1$ of its sentences. The same is done for each of the $K-1$ non-dominant topics, taking $m_2$ sentences from an article with at least $m_2$ of them. Sentences are drawn without replacement within an article and keep their original order; source articles are reused across documents only when the pool is smaller than $n$. The drawn sentences are finally concatenated by topic, so a document consists of $K$ contiguous blocks.
    \item Data preprocessing is applied on the resulting corpus as described
    in~\Cref{app:preprocessing}.
\end{enumerate}
In~\Cref{tab:bbc_detail,tab:20k_detail,tab:wiki_detail} we provide the topic
compositions of the synthetic corpora for each of the runs.
\begin{table}[htbp!]
\centering
\caption{\textbf{BBC News.} For each number of topics $K$ and each repetition, the classes were sampled from BBC News to compose the synthetic documents. Every document draws its dominant topic uniformly among the $K$ listed. When $K = K_{\max}$, a single subset exists, so repetitions vary the corpus seed instead of the classes.}
\begin{tabularx}{\linewidth}{cccX}
\toprule
$K$ & Run & Seed & Topics \\
\midrule
2 & 0 & 6142 & entertainment, politics \\
2 & 1 & 6142 & business, entertainment \\
2 & 2 & 6142 & politics, tech \\
\midrule
3 & 0 & 6142 & entertainment, politics, tech \\
3 & 1 & 6142 & business, entertainment, tech \\
3 & 2 & 6142 & entertainment, sport, tech \\
\midrule
5 & 0 & 6142 & business, entertainment, politics, sport, tech \\
5 & 1 & 6143 & business, entertainment, politics, sport, tech \\
5 & 2 & 6144 & business, entertainment, politics, sport, tech \\
\bottomrule
\end{tabularx}
\label{tab:bbc_detail}
\end{table}

\begin{table}[htbp!]
\centering
\caption{\textbf{20 Newsgroups.} For each number of topics $K$ and each repetition, the classes were sampled from 20 Newsgroups to compose the synthetic documents. Every document draws its dominant topic uniformly among the $K$ listed. When $K = K_{\max}$, a single subset exists, so repetitions vary the corpus seed instead of the classes.}
\begin{tabularx}{\linewidth}{cccX}
\toprule
$K$ & Run & Seed & Topics \\
\midrule
2 & 0 & 6142 & rec.motorcycles, sci.crypt \\
2 & 1 & 6142 & misc.forsale, sci.crypt \\
2 & 2 & 6142 & alt.atheism, rec.autos \\
\midrule
3 & 0 & 6142 & rec.autos, sci.crypt, talk.religion.misc \\
3 & 1 & 6142 & alt.atheism, comp.sys.mac.hardware, rec.autos \\
3 & 2 & 6142 & rec.motorcycles, soc.religion.christian, talk.politics.mideast \\
\midrule
5 & 0 & 6142 & misc.forsale, rec.autos, rec.sport.baseball, sci.crypt, talk.politics.mideast \\
5 & 1 & 6142 & rec.autos, rec.motorcycles, rec.sport.baseball, soc.religion.christian, talk.politics.guns \\
5 & 2 & 6142 & comp.os.ms-windows.misc, misc.forsale, sci.space, talk.politics.guns, talk.politics.misc \\
\midrule
9 & 0 & 6142 & alt.atheism, comp.sys.ibm.pc.hardware, comp.windows.x, rec.autos, rec.sport.baseball, sci.med, sci.space, soc.religion.christian, talk.politics.mideast \\
9 & 1 & 6142 & comp.graphics, comp.sys.mac.hardware, comp.windows.x, rec.motorcycles, rec.sport.hockey, sci.crypt, sci.med, soc.religion.christian, talk.politics.misc \\
9 & 2 & 6142 & alt.atheism, comp.os.ms-windows.misc, comp.sys.mac.hardware, misc.forsale, rec.motorcycles, sci.med, talk.politics.guns, talk.politics.mideast, talk.politics.misc \\
\bottomrule
\end{tabularx}
\label{tab:20k_detail}
\end{table}

\begin{table}[htbp!]
\centering
\caption{\textbf{Wikipedia.} For each number of topics $K$ and each repetition, the classes were sampled from Wikipedia to compose the synthetic documents. Every document draws its dominant topic uniformly among the $K$ listed. When $K = K_{\max}$, a single subset exists, so repetitions vary the corpus seed instead of the classes.}
\begin{tabularx}{\linewidth}{cccX}
\toprule
$K$ & Run & Seed & Topics \\
\midrule
2 & 0 & 6142 & Mathematics; Music \\
2 & 1 & 6142 & Language and literature; Music \\
2 & 2 & 6142 & Agriculture, food, and drink; Language and literature \\
\midrule
3 & 0 & 6142 & Language and literature; Music; Warfare \\
3 & 1 & 6142 & Agriculture, food, and drink; Engineering and technology; Language and literature \\
3 & 2 & 6142 & Language and literature; Social sciences and society; Video games \\
\midrule
5 & 0 & 6142 & History; Language and literature; Mathematics; Music; Sports and recreation \\
5 & 1 & 6142 & Language and literature; Media and drama; Philosophy and religion; Social sciences and society; Sports and recreation \\
5 & 2 & 6142 & Art and architecture; History; Philosophy and religion; Social sciences and society; Warfare \\
\midrule
9 & 0 & 6142 & Agriculture, food, and drink; Engineering and technology; Geography and places; History; Language and literature; Mathematics; Music; Natural sciences; Social sciences and society \\
9 & 1 & 6142 & Art and architecture; Geography and places; Language and literature; Mathematics; Media and drama; Music; Social sciences and society; Video games; Warfare \\
9 & 2 & 6142 & Agriculture, food, and drink; Art and architecture; Geography and places; History; Language and literature; Mathematics; Music; Sports and recreation; Warfare \\
\bottomrule
\end{tabularx}
\label{tab:wiki_detail}
\end{table}

\subsubsection{Preprocessing}
\label{app:preprocessing}
Each document is represented as \textit{(i)} a bag-of-words over a fixed vocabulary and \textit{(ii)} a sequence of sentences.

\paragraph{Lexical view.}
We follow standard ETM preprocessing~\citep{dieng2020topic}. Documents are lemmatized with \texttt{en\_core\_web\_lg}; NLTK stop words, generic verbs and modals, punctuation, non-alphabetic tokens, and tokens shorter than three characters are removed. We also discard terms occurring in more than $\tau_{\max}=0.2$ of the documents or in only one document. The resulting lemmas are intersected with the lowercase alphabetic \texttt{GloVe} vocabulary, retaining the $V=15000$ most frequent words. Thus, all columns of $\brho\in\mathbb R^{p\times V}$ ($p=300$) are initialized with pre-trained \texttt{GloVe} vectors and fine-tuned during training. Documents with fewer than five in-vocabulary tokens are discarded, and counts are normalized before being fed to the inference network.

\paragraph{Contextual view.}
Raw documents are segmented with the \texttt{Punkt} tokenizer as
$\mathbf w^{(d)}=(\mathbf w^{(d)}_1,\dots,\mathbf w^{(d)}_{M^{(d)}})$; if segmentation returns no sentence, the whole document is used as one sentence. Each sentence has a discrete bag-of-words representation over the vocabulary above, used by the lexical evidence, and a continuous representation obtained by encoding the raw sentence with a pre-trained encoder (\texttt{all-mpnet-base-v2} and \texttt{all-MiniLM-L6-v2} for the synthetic and real corpora, and \texttt{Qwen3-Embedding-0.6B} for the debate). The resulting embedding $\mathbf e^{(d)}_m$ is $\ell_2$-normalized, $\tilde{\mathbf e}^{(d)}_m=\mathbf e^{(d)}_m/\|\mathbf e^{(d)}_m\|$, and used as contextual evidence (\Cref{sec:method}). Embeddings are computed once and cached; the encoders are not fine-tuned.

\subsection{ElecDeb60to20}
The organization of the debate is indicated by the journalist at the beginning of the debate:

``\textit{Good evening from the Health Education Campus of Case Western Reserve University and the Cleveland Clinic. I’m Chris Wallace of Fox News and I welcome you to the first of the 2020 presidential debates between President Donald J. Trump and former Vice President Joe Biden. This debate is sponsored by the Commission on Presidential Debates. The Commission has designed the format, six roughly 15-minute segments with two-minute answers from each candidate to the first question, then open discussion for the rest of each segment. Both campaigns have agreed to these rules. For the record, I decided the topics and the questions in each topic. I can assure you none of the questions has been shared with the Commission or the two candidates. This debate is being conducted under health and safety protocols designed by the Cleveland Clinic, which is serving as the health security advisor to the Commission for all four debates. As a precaution, both campaigns have agreed the candidates will not shake hands at the beginning of tonight’s debate. The audience here in the hall has promised to remain silent. No cheers, no boos, or other interruptions so we, and more importantly you, can focus on what the candidates have to say. No noise except right now, as we welcome the Republican nominee, President Trump, and the Democratic nominee Vice President Biden.}''

Overall, the segments present the following topics: 
\begin{enumerate}
    \item Index 4 -- 146: \texttt{Supreme Court / Healthcare / Obamacare / Filibuster / Court packing}.
    \item Index 147 -- 272: \texttt{COVID19 / Vaccine / Scientists / Reopening schools and economy / Masks / Disease spread at events}.
    \item Index 273 -- 475: \texttt{Economy / Unemployment rate / Recovery / Taxes / Economic plan}.
    \item Index 476 -- 626: \texttt{Race issues / Racial sensitivity training / Law and Order / Black Lives Matter / Protests}.
    \item Index 627 -- 779: \texttt{Trump and Biden records / Climate change / Global warming / Green New Deal}.
    \item Index 780 -- 853: \texttt{Election integrity / Mail-in voting}.
\end{enumerate}

\subsection{Model training}
Sentence embeddings are computed once with a frozen encoder: \texttt{all-mpnet-base-v2} in the main experiments and \texttt{all-MiniLM-L6-v2} in the encoder comparison. For models using word embeddings, $\brho$ is initialized with \texttt{GloVe} and fine-tuned.
{\sentetm} is trained for 300 epochs with Adam (learning rate $2\cdot10^{-3}$, weight decay $1.2\cdot10^{-6}$, gradient clipping $2.0$) and batches of 64 documents. The inference network has hidden size 256, topic embeddings have dimension 300, the topic subspace dimension is (for all the topics) $h_k=32$, and contextual evidence is weighted by $c=0.5$. Topic Gaussians are initialized by k-means (10 restarts) on normalized sentence embeddings and re-estimated in closed form after each epoch, rather than optimized by Adam. Eigenvalues and noise variances are lower-bounded by $10^{-6}$.
For the baselines, we use the hyperparameters from the original implementations, sharing the batch size and learning rate across neural models. ETM~\citep{dieng2020topic} is trained for 150 epochs with hidden size 800, no encoder dropout, and weight decay $1.2\cdot10^{-6}$. LDA~\citep{blei2003latent} uses 201 batch iterations; senLDA~\citep{balikas2016topic}, 201 Gibbs sweeps with $\alpha=\beta=1/K$; and SentenceLDA~\citep{cha-lee-2024-sentencelda}, 20 sweeps with $\alpha=\kappa=0.1$. For the Gibbs samplers, we use the final state without averaging, as in the original implementations. BERTopic~\citep{grootendorst2022bertopic} uses UMAP with 15 neighbors, 5 components, and cosine distance, is constrained to $K$ topics, and reassigns outliers by embedding similarity.
Each configuration is repeated 3 times, varying source classes or, when only one class subset is available, the sampled documents and sentences. Each model is trained with 3 seeds per repetition. 

\subsection{Metrics} 
We measure assignment quality against the ground-truth sentence topics with the Adjusted Mutual Information~\citep{vinh2009information} and the Adjusted Rand Index~\citep{hubert1985comparing}, both corrected for chance. Since every sentence has a known topic, these metrics directly evaluate the quantity that the models infer, without resorting to proxies for topic quality.

\subsection{Computational cost}
\label{app:cost}
The cost of {\sentetm} has three parts. Let $S=\sum_d M^{(d)}$ be the number of sentences in the corpus.

\textit{(i) Encoding.} Each sentence is encoded once by the frozen encoder, and the embeddings are cached. This cost is shared with every method that relies on sentence embeddings; with \texttt{all-mpnet-base-v2}, encoding the $14{,}000$ sentences of a synthetic corpus takes about 30 seconds ($\approx$2\, ms per sentence).

\textit{(ii) Gradient steps.} At each epoch, the inference network costs $O(DVH)$, with $H$ its hidden size; the lexical evidence costs $O(KVp)$ to compute $\bm\beta$ plus one term per word token and topic; the contextual evidence costs $O(SE\sum_{k=1}^K h_k)$, which reduces to $O(SKEh)$ when $h_k=h$ for all $k$.

\textit{(iii) Closed-form update.} Once per epoch, the topic Gaussians are re-estimated from the responsibilities, which requires $O(SKE^2)$ operations to accumulate the covariances and $O(KE^3)$ for their eigendecompositions. This step does not depend on the vocabulary and grows linearly with $S$ and $K$.

\Cref{tab:cost} reports the training time of all models on all the datasets. \textbf{On average over $K$, {\sentetm} trains in 340 to 391 seconds per run, against 1717 to 3118 seconds for SentenceLDA, the other sentence-level model that uses sentence embeddings whose Gibbs sampler must update every sentence assignment sequentially}. It is slower than the document-level models and senLDA, which use neither embeddings nor a per-sentence posterior over $K$ Gaussians. Its cost grows linearly with $K$. \textbf{Most of this time is spent in the closed-form update, which our implementation runs on the CPU; it could be moved to the GPU or performed every few epochs.} Once the model is trained, assigning a topic to a new sentence requires one forward pass of the inference network and the evaluation of $K$ Gaussian densities.
\begin{table}[!h]
\centering
\caption{Training time in seconds (median over 9 runs; 700 documents, $14{,}000$ sentences), excluding corpus construction and sentence encoding. All experiments were run on NVIDIA H100 GPUs. Neural models (SentenceETM, ETM, BERTopic, SenClu) run on a GPU; LDA, senLDA, and SentenceLDA, whose released implementations are CPU-only, run on CPU.}
\label{tab:cost}
\begin{tabular}{llrrrrrrr}
\toprule
 & & \multicolumn{4}{c}{Sentence-level} & \multicolumn{3}{c}{Document-level} \\
\cmidrule(lr){3-6}\cmidrule(lr){7-9}
Dataset & $K$ & {\sentetm} & SentenceLDA & senLDA & SenClu & BERTopic & ETM & LDA \\
\midrule
\multirow{3}{*}{BBC News} & 2 & 281 & 1946 & 86 & 1.6 & 1.6 & 4.3 & 26 \\
 & 3 & 328 & 1391 & 87 & 1.9 & 1.6 & 4.4 & 23 \\
 & 5 & 410 & 1815 & 88 & 2.3 & 1.7 & 4.6 & 24 \\
\cmidrule(lr){2-9}
 & \textit{Avg.} & 340 & 1717 & 87 & 1.9 & 1.6 & 4.4 & 25 \\
\midrule
\multirow{4}{*}{20 Newsgroups} & 2 & 210 & 1913 & 81 & 1.2 & 1.5 & 4.0 & 22 \\
 & 3 & 255 & 2116 & 81 & 1.9 & 1.5 & 4.1 & 21 \\
 & 5 & 335 & 1817 & 82 & 2.0 & 1.5 & 4.1 & 20 \\
 & 9 & 497 & 5309 & 82 & 2.7 & 1.5 & 4.2 & 19 \\
\cmidrule(lr){2-9}
 & \textit{Avg.} & 324 & 2789 & 82 & 2.0 & 1.5 & 4.1 & 21 \\
\midrule
\multirow{4}{*}{Wikipedia} & 2 & 268 & 2308 & 85 & 1.8 & 1.5 & 3.7 & 26 \\
 & 3 & 328 & 2304 & 84 & 1.8 & 1.5 & 3.8 & 24 \\
 & 5 & 413 & 2684 & 84 & 2.2 & 1.5 & 3.9 & 23 \\
 & 9 & 554 & 5176 & 88 & 2.2 & 1.4 & 4.1 & 23 \\
\cmidrule(lr){2-9}
 & \textit{Avg.} & 391 & 3118 & 85 & 2.0 & 1.5 & 3.9 & 24 \\
\bottomrule
\end{tabular}
\end{table}
\section{Additional Materials to Section~\ref{sec:exp_results}}
\label{app:exp_results}
\subsection{Numerical results}
\label{app:senclu}
\Cref{tab:results_bbc,tab:results_20k,tab:results_wiki} report its ARI and AMI on the synthetic corpora. The tables also present the results for SenClu~\citep{schneider2023efficient}, which assigns sentences to topics by hard expectation-maximization over sentence embeddings, weighted by document-level topic proportions.
We use the provided official implementation with its default hyperparameters (40 epochs, $\alpha=1/\sqrt{K}$) and feed it the same sentences and sentence embeddings as the other models. {\sentetm} outperforms it in every setting. SenClu varies strongly across runs (standard deviation up to 0.32 ARI on Wikipedia), which we attribute to its random initialization of the topic centroids.

\subsection{Structure of the latent space}
\label{app:latent}
\begin{figure}[!t]
\centering
    \begin{subfigure}{.49\columnwidth}
    \centering
    \includegraphics[width=\columnwidth]{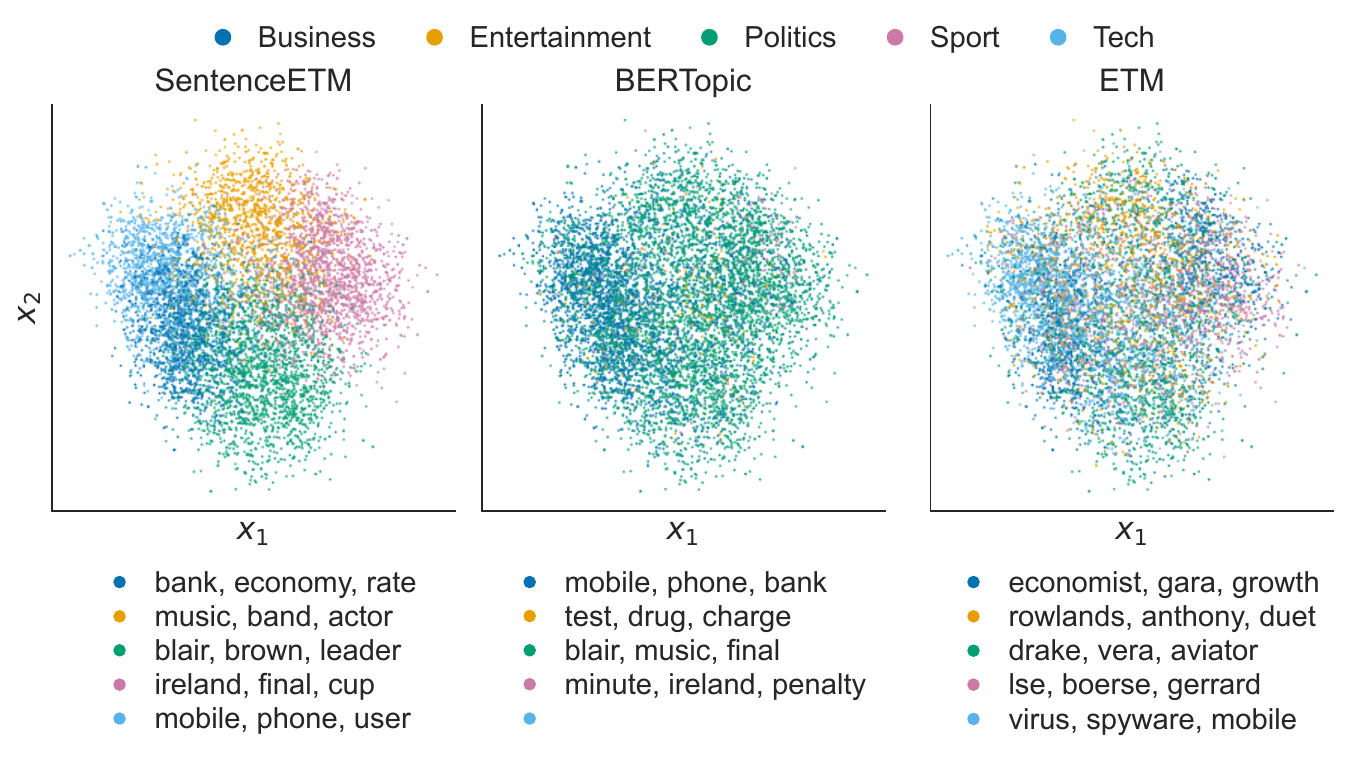}
    \caption{BBC News}
    \label{fig:embeddings_bbc}
\end{subfigure}
\hfill
    \begin{subfigure}{.49\columnwidth}
    \centering
    \includegraphics[width=\columnwidth]{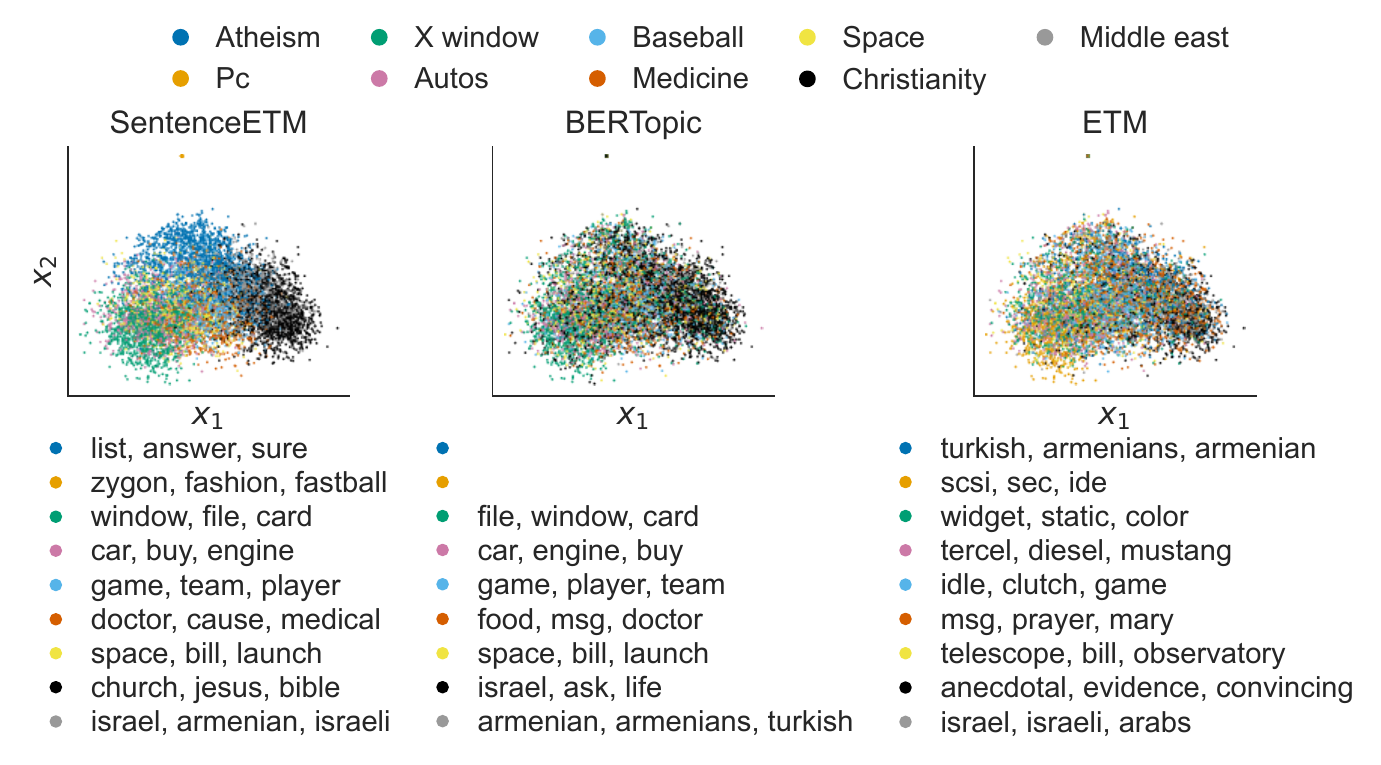}
    \caption{20 Newsgroups}
    \label{fig:embeddings_20k}
\end{subfigure}

\caption{\textbf{Sentence-level topic assignments} (\texttt{all-mpnet-base-v2}). Sentences projected onto the first two embedding PCs and colored by assigned topic; top-3 words below each panel. Wikipedia is in~\Cref{fig:embeddings_wiki}.}
\label{fig:embeddings}
\end{figure}

\Cref{fig:embeddings,fig:embeddings_wiki} show, for one run per dataset, the sentences projected onto the first two principal components of their embeddings and colored by the topic assigned by each model. The three most probable words of each topic are listed below each panel. With {\sentetm}, sentences assigned to the same topic largely form distinct regions of the embedding space, and the top words identify the corresponding classes (e.g., \textit{bank, economy, rate} for Business in BBC News, \textit{church, jesus, bible} for Christianity in 20 Newsgroups, \textit{music, chart, video} for Music in Wikipedia). In contrast, BERTopic and ETM produce substantially more mixed assignments. BERTopic assigns one topic per document, so every sentence inherits its document's topic regardless of its content; ETM assigns sentence topics from the words of document-level topics. Their top words reveal further limitations: BERTopic produces topics that mix or duplicate classes (e.g., \textit{game, season, team} and \textit{game, league, play} on Wikipedia), while ETM topics are often dominated by rare, specific words (\textit{Ise, boerse, gerrad}; \textit{tercel, diesel, mustang}; \textit{chili, doughnut, steak}). Finally, we notice that {\sentetm} on 20 Newsgroups, the topic matched to Atheism is dominated by conversational words (\textit{list, answer, sure}), while the one matched to X window has uninformative top words and few sentences, consistent with the lower scores on this dataset. In Appendix~\ref{app:misassigned} we provide some examples of the misassigned sentences.

\subsection{Sensitivity to the sentence encoder}
\label{app:encoder}
Replacing \texttt{all-mpnet-base-v2} with the smaller \texttt{all-MiniLM-L6-v2} affects the embedding-based sentence models very differently. For {\sentetm}, the drop is at most 0.05 in both ARI and AMI, except on BBC News at $K=5$ (0.19 ARI and 0.15 AMI), where its ARI falls slightly below that of senLDA. SentenceLDA, in contrast, degrades sharply: its ARI falls to at most 0.16 in every setting (e.g., from 0.52 to 0.02 on BBC News at $K=2$), and its AMI decreases by up to 0.39. With \texttt{all-MiniLM-L6-v2}, SentenceLDA obtains lower ARI than senLDA, which does not use embeddings, in every setting. This degradation cannot be attributed solely to the weaker embeddings: SenClu, using the same \texttt{all-MiniLM-L6-v2} embeddings, loses at most 0.06 ARI relative to \texttt{all-mpnet-base-v2} and even improves in most settings. A possible explanation lies in how the models represent topics in the embedding space. SentenceLDA uses full-covariance Gaussians, making its clustering depend on the geometry of the entire space. In {\sentetm}, each covariance is instead constrained to a few principal directions plus isotropic noise and re-estimated from soft responsibilities; this parsimonious representation may be less sensitive to directions carrying little topical information. Note that senLDA, ETM, and LDA do not use the sentence encoder, so their results are identical for both encoders (cf.~\Cref{tab:results_bbc,tab:results_20k,tab:results_wiki}).
\newcommand{\sd}[1]{{\scriptsize$\pm$#1}}
\begin{table*}[!t]
\caption{\textbf{BBC News: ARI and AMI.} Results are reported as mean $\pm$ standard deviation over $3 \times 3$ runs. \textbf{Bold} the best results within each sentence encoder; \underline{underlined} the overall best results.}
\centering
\setlength{\tabcolsep}{3pt}
\begin{tabular}{cl ccc ccc}
\toprule
& & \multicolumn{3}{c}{ARI} & \multicolumn{3}{c}{AMI} \\
\cmidrule(lr){3-5}\cmidrule(lr){6-8}
 & Model & $K=2$ & $K=3$ & $K=5$ & $K=2$ & $K=3$ & $K=5$ \\
\midrule
\multirow{6}{*}{\rotatebox{90}{\texttt{mpnet}}}
& SentenceETM &  \underline{\textbf{0.76}} \sd{0.04}  &  \underline{\textbf{0.69}} \sd{0.03}  &  \underline{\textbf{0.63}} \sd{0.01} &  \underline{\textbf{0.67}} \sd{0.04}  &  \underline{\textbf{0.62}} \sd{0.03}  &  \underline{\textbf{0.61}} \sd{0.01}\\
& SentenceLDA &  0.52 \sd{0.12}  &  0.58 \sd{0.03}  &  0.54 \sd{0.07} &  0.50 \sd{0.08}  &  0.54 \sd{0.02}  &  0.52 \sd{0.05}\\
& senLDA &  0.52 \sd{0.08}  &  0.43 \sd{0.11}  &  0.46 \sd{0.07} &  0.45 \sd{0.06}  &  0.39 \sd{0.10}  &  0.45 \sd{0.05}\\
& SenClu &  0.26 \sd{0.22}  &  0.25 \sd{0.12}  &  0.23 \sd{0.04}  &  0.25 \sd{0.20}  &  0.26 \sd{0.11}  &  0.26 \sd{0.04}\\
\cmidrule(lr){2-8}
& BERTopic &  0.00 \sd{0.00}  &  0.01 \sd{0.02}  &  0.08 \sd{0.01} &  0.00 \sd{0.00}  &  0.01 \sd{0.01}  &  0.11 \sd{0.01}\\
& ETM &  0.01 \sd{0.01}  &  0.03 \sd{0.02}  &  0.15 \sd{0.04} &  0.01 \sd{0.01}  &  0.03 \sd{0.01}  &  0.15 \sd{0.04}\\
& LDA &  0.00 \sd{0.00}  &  0.00 \sd{0.00}  &  0.02 \sd{0.00} &  0.00 \sd{0.00}  &  0.00 \sd{0.00}  &  0.02 \sd{0.00}\\
\midrule
\midrule
\multirow{6}{*}{\rotatebox{90}{\texttt{MiniLM}}}
& SentenceETM &  \textbf{0.72} \sd{0.04}  &  \textbf{0.64} \sd{0.03}  &  0.44 \sd{0.01} &  \textbf{0.63} \sd{0.04}  &  \textbf{0.58} \sd{0.04}  &  \textbf{0.46} \sd{0.01}\\
& SentenceLDA &  0.02 \sd{0.01}  &  0.04 \sd{0.01}  &  0.12 \sd{0.03} &  0.11 \sd{0.01}  &  0.19 \sd{0.03}  &  0.34 \sd{0.03}\\
& senLDA &  0.52 \sd{0.08}  &  0.43 \sd{0.11}  &  \textbf{0.46} \sd{0.07} &  0.45 \sd{0.06}  &  0.39 \sd{0.10}  &  0.45 \sd{0.05}\\
& SenClu &  0.39 \sd{0.16}  &  0.26 \sd{0.13}  &  0.25 \sd{0.06}  &  0.37 \sd{0.14}  &  0.26 \sd{0.13}  &  0.27 \sd{0.04}\\
\cmidrule(lr){2-8}
& BERTopic &  0.00 \sd{0.00}  &  0.03 \sd{0.02}  &  0.10 \sd{0.04} &  0.00 \sd{0.00}  &  0.03 \sd{0.02}  &  0.12 \sd{0.03}\\
& ETM &  0.01 \sd{0.01}  &  0.03 \sd{0.02}  &  0.15 \sd{0.04} &  0.01 \sd{0.01}  &  0.03 \sd{0.01}  &  0.15 \sd{0.04}\\
& LDA &  0.00 \sd{0.00}  &  0.00 \sd{0.00}  &  0.02 \sd{0.00} &  0.00 \sd{0.00}  &  0.00 \sd{0.00}  &  0.02 \sd{0.00}\\
\bottomrule
\end{tabular}
\label{tab:results_bbc}
\end{table*}

\begin{table*}[!htbp]
\centering
\caption{\textbf{20 Newsgroup: ARI and AMI}. Results are reported as mean $\pm$ standard deviation over $3 \times 3$ runs. \textbf{Bold} the best results within each sentence encoder; \underline{underlined} the overall best results.}
\setlength{\tabcolsep}{3pt}
\begin{tabular}{cl cccc cccc}
\toprule
& & \multicolumn{4}{c}{ARI} & \multicolumn{4}{c}{AMI} \\
\cmidrule(lr){3-6}\cmidrule(lr){7-10}
 & Model & $K=2$ & $K=3$ & $K=5$ & $K=9$ & $K=2$ & $K=3$ & $K=5$ & $K=9$ \\
\midrule
\multirow{6}{*}{\rotatebox{90}{\texttt{mpnet}}}
& SentenceETM &  \underline{\textbf{0.56}} \sd{0.05}  &  \underline{\textbf{0.46}} \sd{0.03}  &  \underline{\textbf{0.35}} \sd{0.04}  &  \underline{\textbf{0.27}} \sd{0.02} &  \underline{\textbf{0.48}} \sd{0.03}  &  \underline{\textbf{0.44}} \sd{0.02}  &  \underline{\textbf{0.41}} \sd{0.05}  &  \underline{\textbf{0.37}} \sd{0.04}\\
& SentenceLDA &  0.16 \sd{0.04}  &  0.23 \sd{0.02}  &  0.30 \sd{0.07}  &  0.25 \sd{0.03} &  0.26 \sd{0.03}  &  0.37 \sd{0.01}  &  0.39 \sd{0.05}  &  0.33 \sd{0.03}\\
& senLDA &  0.22 \sd{0.13}  &  0.30 \sd{0.02}  &  0.21 \sd{0.03}  &  0.17 \sd{0.05} &  0.18 \sd{0.08}  &  0.26 \sd{0.01}  &  0.22 \sd{0.03}  &  0.21 \sd{0.05}\\
& SenClu &  0.36 \sd{0.19}  &  0.35 \sd{0.07}  &  0.28 \sd{0.03}  &  0.23 \sd{0.04}  &  0.32 \sd{0.16}  &  0.33 \sd{0.06}  &  0.33 \sd{0.03}  &  0.31 \sd{0.05}\\
\cmidrule(lr){2-10}
& BERTopic &  0.00 \sd{0.00}  &  0.02 \sd{0.03}  &  0.09 \sd{0.02}  &  0.06 \sd{0.08} &  0.00 \sd{0.00}  &  0.03 \sd{0.03}  &  0.12 \sd{0.02}  &  0.10 \sd{0.10}\\
& ETM &  0.02 \sd{0.02}  &  0.12 \sd{0.02}  &  0.11 \sd{0.01}  &  0.11 \sd{0.03} &  0.02 \sd{0.02}  &  0.11 \sd{0.02}  &  0.12 \sd{0.02}  &  0.15 \sd{0.04}\\
& LDA &  0.00 \sd{0.00}  &  0.01 \sd{0.01}  &  0.01 \sd{0.00}  &  0.02 \sd{0.01} &  0.00 \sd{0.00}  &  0.01 \sd{0.01}  &  0.02 \sd{0.01}  &  0.03 \sd{0.01}\\
\midrule
\midrule
\multirow{6}{*}{\rotatebox{90}{\texttt{MiniLM}}}
& SentenceETM &  \textbf{0.54} \sd{0.04}  &  \textbf{0.43} \sd{0.02}  &  \textbf{0.32} \sd{0.01}  &  \textbf{0.25} \sd{0.03} &  \textbf{0.46} \sd{0.02}  &  \textbf{0.41} \sd{0.02}  &  \textbf{0.36} \sd{0.03}  &  \textbf{0.34} \sd{0.04}\\
& SentenceLDA &  0.01 \sd{0.00}  &  0.02 \sd{0.01}  &  0.03 \sd{0.01}  &  0.03 \sd{0.01} &  0.07 \sd{0.00}  &  0.13 \sd{0.02}  &  0.22 \sd{0.03}  &  0.25 \sd{0.04}\\
& senLDA &  0.22 \sd{0.13}  &  0.30 \sd{0.02}  &  0.21 \sd{0.03}  &  0.17 \sd{0.05} &  0.18 \sd{0.08}  &  0.26 \sd{0.01}  &  0.22 \sd{0.03}  &  0.21 \sd{0.05}\\
& SenClu &  0.40 \sd{0.08}  &  0.29 \sd{0.04}  &  0.25 \sd{0.05}  &  0.20 \sd{0.04}  &  0.33 \sd{0.07}  &  0.28 \sd{0.02}  &  0.28 \sd{0.04}  &  0.28 \sd{0.04}\\ 
\cmidrule(lr){2-10}
& BERTopic &  0.00 \sd{0.00}  &  0.02 \sd{0.03}  &  0.07 \sd{0.06}  &  0.06 \sd{0.08} &  0.00 \sd{0.00}  &  0.03 \sd{0.03}  &  0.09 \sd{0.07}  &  0.10 \sd{0.09}\\
& ETM &  0.02 \sd{0.02}  &  0.12 \sd{0.02}  &  0.11 \sd{0.01}  &  0.11 \sd{0.03} &  0.02 \sd{0.02}  &  0.11 \sd{0.02}  &  0.12 \sd{0.02}  &  0.15 \sd{0.04}\\
& LDA &  0.00 \sd{0.00}  &  0.01 \sd{0.01}  &  0.01 \sd{0.00}  &  0.02 \sd{0.01} &  0.00 \sd{0.00}  &  0.01 \sd{0.01}  &  0.02 \sd{0.01}  &  0.03 \sd{0.01}\\

\bottomrule
\end{tabular}
\label{tab:results_20k}
\end{table*}

\begin{table}[!htbp]
\centering
\caption{\textbf{Wikipedia: ARI and AMI.} Results are reported as mean $\pm$ standard deviation over $3 \times 3$ runs. \textbf{Bold} the best results within each sentence encoder; \underline{underlined} the overall best results.}
\setlength{\tabcolsep}{3pt}
\begin{tabular}{cl cccc cccc}
\toprule
& & \multicolumn{4}{c}{ARI} & \multicolumn{4}{c}{AMI} \\
\cmidrule(lr){3-6}\cmidrule(lr){7-10}
 & Model & $K=2$ & $K=3$ & $K=5$ & $K=9$ & $K=2$ & $K=3$ & $K=5$ & $K=9$ \\
\midrule
\multirow{6}{*}{\rotatebox{90}{\texttt{mpnet}}}
& SentenceETM &  \underline{\textbf{0.69}} \sd{0.13}  &  \underline{\textbf{0.57}} \sd{0.08}  &  \underline{\textbf{0.45}} \sd{0.15}  &  \underline{\textbf{0.47}} \sd{0.04} &  \underline{\textbf{0.61}} \sd{0.15}  &  \underline{\textbf{0.55}} \sd{0.09}  &  \underline{\textbf{0.47}} \sd{0.15}  &  \underline{\textbf{0.55}} \sd{0.02}\\
& SentenceLDA &  0.44 \sd{0.11}  &  0.41 \sd{0.07}  &  0.38 \sd{0.15}  &  0.45 \sd{0.05} &  0.45 \sd{0.09}  &  0.48 \sd{0.09}  &  0.42 \sd{0.15}  &  0.50 \sd{0.04}\\
& senLDA &  0.53 \sd{0.19}  &  0.47 \sd{0.08}  &  0.31 \sd{0.09}  &  0.29 \sd{0.04} &  0.46 \sd{0.18}  &  0.43 \sd{0.08}  &  0.34 \sd{0.09}  &  0.38 \sd{0.03}\\
& SenClu &  0.37 \sd{0.32}  &  0.40 \sd{0.16}  &  0.22 \sd{0.15}  &  0.27 \sd{0.04}  &  0.34 \sd{0.28}  &  0.38 \sd{0.15}  &  0.23 \sd{0.15}  &  0.34 \sd{0.03}\\
\cmidrule(lr){2-10}
& BERTopic &  0.00 \sd{0.00}  &  0.00 \sd{0.00}  &  0.04 \sd{0.02}  &  0.11 \sd{0.05} &  0.00 \sd{0.00}  &  0.00 \sd{0.00}  &  0.08 \sd{0.01}  &  0.18 \sd{0.03}\\
& ETM &  0.02 \sd{0.02}  &  0.17 \sd{0.06}  &  0.14 \sd{0.07}  &  0.20 \sd{0.02} &  0.02 \sd{0.02}  &  0.16 \sd{0.05}  &  0.15 \sd{0.06}  &  0.24 \sd{0.01}\\
& LDA &  0.00 \sd{0.00}  &  0.01 \sd{0.01}  &  0.02 \sd{0.01}  &  0.03 \sd{0.01} &  0.00 \sd{0.00}  &  0.01 \sd{0.01}  &  0.03 \sd{0.02}  &  0.05 \sd{0.01}\\

\midrule
\midrule

\multirow{6}{*}{\rotatebox{90}{\texttt{MiniLM}}}
& SentenceETM &   \textbf{0.67} \sd{0.14}  &  \underline{\textbf{0.57}} \sd{0.10}  &  \textbf{0.43} \sd{0.15}  & \textbf{ 0.44} \sd{0.05} &  \textbf{0.60} \sd{0.15}  &  \textbf{0.53} \sd{0.11}  &  \textbf{0.45} \sd{0.14}  &  \textbf{0.51} \sd{0.02}\\
& SentenceLDA &  0.03 \sd{0.02}  &  0.10 \sd{0.05}  &  0.11 \sd{0.01}  &  0.16 \sd{0.04} &  0.11 \sd{0.03}  &  0.27 \sd{0.06}  &  0.29 \sd{0.05}  &  0.38 \sd{0.03}\\
& senLDA &  0.53 \sd{0.19}  &  0.47 \sd{0.08}  &  0.31 \sd{0.09}  &  0.29 \sd{0.04} &  0.46 \sd{0.18}  &  0.43 \sd{0.08}  &  0.34 \sd{0.09}  &  0.38 \sd{0.03}\\
& SenClu &  0.52 \sd{0.30}  &  0.46 \sd{0.18}  &  0.27 \sd{0.15}  &  0.31 \sd{0.03}  &  0.46 \sd{0.28}  &  0.42 \sd{0.16}  &  0.28 \sd{0.15}  &  0.38 \sd{0.02}\\
\cmidrule(lr){2-10}
& BERTopic &  0.00 \sd{0.00}  &  0.00 \sd{0.00}  &  0.03 \sd{0.01}  &  0.13 \sd{0.04} &  0.00 \sd{0.00}  &  0.00 \sd{0.00}  &  0.08 \sd{0.01}  &  0.19 \sd{0.03}\\
& ETM &  0.02 \sd{0.02}  &  0.17 \sd{0.06}  &  0.14 \sd{0.07}  &  0.20 \sd{0.02} &  0.02 \sd{0.02}  &  0.16 \sd{0.05}  &  0.15 \sd{0.06}  &  0.24 \sd{0.01}\\
& LDA &  0.00 \sd{0.00}  &  0.01 \sd{0.01}  &  0.02 \sd{0.01}  &  0.03 \sd{0.01} &  0.00 \sd{0.00}  &  0.01 \sd{0.01}  &  0.03 \sd{0.02}  &  0.05 \sd{0.01}\\
\bottomrule
\end{tabular}
\label{tab:results_wiki}
\end{table}

\subsection{Sensitivity to document length}
\label{app:doc_length}
\begin{figure}[!ht]
    \centering
    \includegraphics[width=\columnwidth]{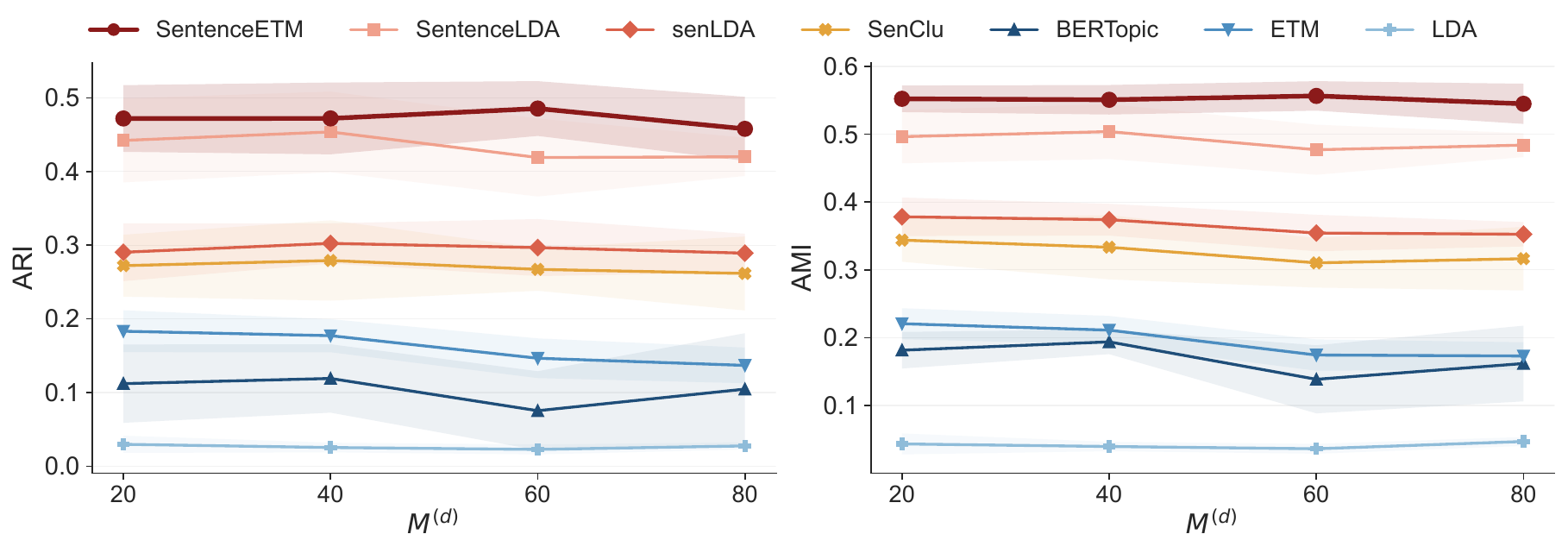}
    \caption{\textbf{Sensitivity to document length} (Wikipedia, $K=9$, \texttt{all-mpnet-base-v2}). 
ARI (left) and AMI (right) as a function of the number of sentences per document $M^{(d)}$ ($3\times 3$ runs).}
    \label{fig:doc_length}
\end{figure}
We vary the number of sentences per document $M^{(d)}\in\{20,40,60,80\}$ on Wikipedia with $K=9$ and \texttt{all-mpnet-base-v2}, while decreasing the number of documents accordingly (700, 350, 233, and 175), so that each corpus contains approximately $14{,}000$ sentences. All other settings are the same as in the main experiments.

As shown in~\Cref{fig:doc_length}, {\sentetm} obtains the highest ARI and AMI at every document length, while remaining stable across the four settings: ARI ranges from 0.46 to 0.49 and AMI from 0.54 to 0.56, with differences within one standard deviation. The other sentence-level models are similarly stable but consistently lower, with SentenceLDA ranging from 0.42 to 0.45 ARI and senLDA remaining around 0.30. Among the document-level models, ETM gradually degrades as document length increases (ARI from 0.18 to 0.14), BERTopic fluctuates between 0.08 and 0.12, and LDA remains close to zero. These results suggest that sentence-level assignments are largely insensitive to document length in this setting, even when the same amount of sentence-level information is distributed across fewer, longer documents.
\subsection{Results on original corpora}
\label{app:original}
\begin{figure}[!ht]
    \centering
    \includegraphics[width=.75\columnwidth]{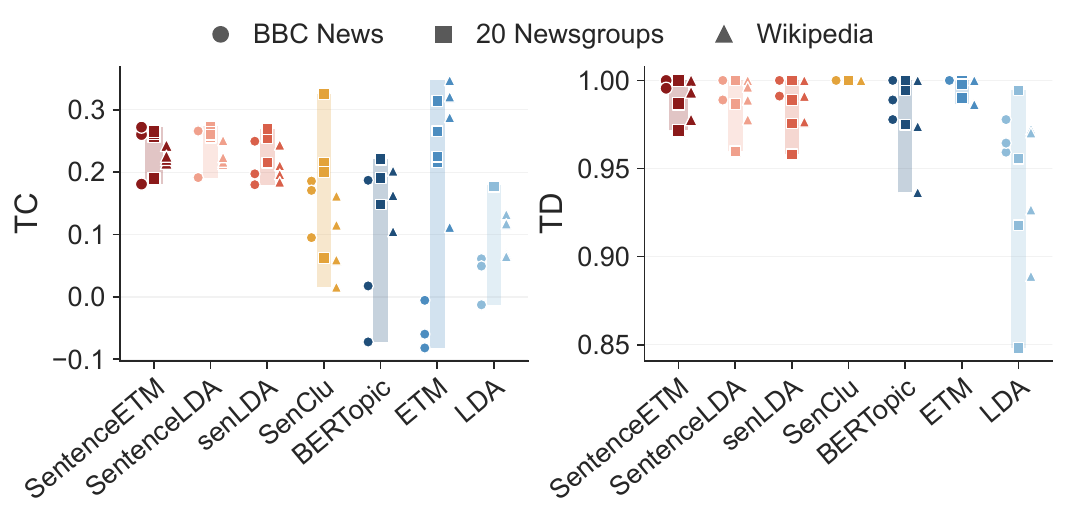}
\caption{\textbf{Topic coherence (TC) and topic diversity (TD) on the original corpora.}
Topics are those learned in the synthetic-corpus experiments, while TC is evaluated using word co-occurrences from the corresponding full original dataset. TD is computed directly from the learned top words. Higher is better.}   
\label{fig:tc_td}
\end{figure}
We evaluate the lexical quality of the learned topics (10 top-words) through Topic Coherence (TC) with NPMI~\citep{mimno2011optimizing,lau2014machine}   and Topic Diversity (TD)~\citep{dieng2020topic}. We use the topics learned in the same runs as the synthetic-corpus experiments; no model is retrained on the original datasets. For each run, we extract the top words of the learned topics and evaluate them against the corresponding full original corpus. This choice avoids measuring coherence on the synthetic documents used for training, whose co-occurrence statistics are partly induced by our construction procedure: each synthetic document deliberately combines sentences drawn from different source classes, and words from otherwise unrelated topics may therefore co-occur simply because their sentences were placed in the same document. 
Computing TC on these documents could consequently reward topic-word sets that reflect the synthetic mixing procedure rather than genuine associations in the source corpus. 
Evaluating the same learned top words on the original corpus instead measures whether they correspond to word associations that occur naturally in the underlying data.
TD does not require a reference corpus, since it depends only on the overlap among the top words of the learned topics. 
We therefore compute it directly from the topics obtained in each run.~\Cref {fig:tc_td} shows that the sentence-level models produce consistently coherent and diverse topics across datasets. {\sentetm} maintains high TC on BBC News, 20 Newsgroups, and Wikipedia, with comparatively limited variation across datasets. SentenceLDA and senLDA show a similarly stable behavior, whereas the document-level baselines are substantially more dataset-dependent: in particular, ETM ranges from low or negative coherence on BBC News and 20 Newsgroups to among the highest values on Wikipedia. BERTopic and LDA exhibit similar, although less pronounced, variations. TD is close to one for most methods and configurations, especially for the sentence-level models, indicating little overlap among their top words. It therefore provides less discrimination than TC in this setting; importantly, {\sentetm} combines this high diversity with consistently high coherence.

\subsection{Examples of misassigned sentences}
\label{app:misassigned}
In~\Cref{tab:misassignements}, we randomly select eight sentences assigned to an incorrect topic by {\sentetm} for each dataset, using the same runs as in~\Cref{fig:embeddings,fig:embeddings_wiki}. 
Most errors fall into three groups. \textit{First}, many misassigned sentences contain little topical information: generic statements, reported speech, or discourse-like utterances whose label is inherited from the source document but cannot be reliably inferred from the sentence alone (e.g., \textit{So to win something like this really feels great''}, \textit{``I won't argue too much about anything''}). In such cases, the predicted topic often reflects cues in the wording itself, such as \textit{win} suggesting sport or an argumentative tone suggesting a religion-related forum.
\textit{Second}, several sentences are semantically closer to the predicted topic than to the label of their source document. For example, a business article on Turkish firms excluded for national-security reasons discusses politics; an \texttt{alt.atheism} post mentioning Soviet and French tanks discusses the Middle East conflict; and Wikipedia sentences mentioning boxing or cricket naturally evoke sport. \textbf{These cases highlight the noise introduced by propagating document-level labels to individual sentences}.
\textit{Third}, \textbf{some disagreements involve inherently overlapping categories}, such as \texttt{Warfare} and \texttt{History}, or \texttt{Geography and places} and \texttt{Art and architecture} for a sentence describing a historical house. Only a few cases lack a clear semantic explanation, such as a sentence on the history of Calakmul assigned to \texttt{Language and literature} or a bibliographic reference assigned to \texttt{Geography and places}.

\begin{table}[!thp]
\centering
\caption{\textbf{Examples of topic disagreements with the source labels.} We report the topic inherited by the sentence from its source document and the predicted topic ({\sentetm}). The misassignments are on generic sentences, cases where the predicted topic better matches the sentence content, and overlaps between categories.}
\begin{tabularx}{\textwidth}{r|X}
\toprule
\textbf{True $\rightarrow$ Predicted} & \textbf{Sentence} \\
\midrule

\multicolumn{2}{c}{\textbf{BBC News}} \\[2pt]

\texttt{Entertainment} $\rightarrow$ \texttt{Sport} &
\textit{There were no specific details on the injuries}. \\

\texttt{Business} $\rightarrow$ \texttt{Tech} &
\textit{It will do so at a presentation in New York on 26 January}. \\


\texttt{Business} $\rightarrow$ \texttt{Politics} &
\textit{``I will give up my role as soon as the board of directors asks me to do so,'' he said}. \\

\texttt{Business} $\rightarrow$ \texttt{Politics} &
\textit{In both cases, the justification has been national security, amid allegations that the Turkish firms are too close to Israel}. \\

\texttt{Business} $\rightarrow$ \texttt{Politics} &
\textit{``First and foremost this is a human tragedy.''} \\


\texttt{Tech} $\rightarrow$ \texttt{Business} &
\textit{Taipei's TFC 101 Tower is more than 50m taller than the Petronas Towers in Kuala Lumpur, Malaysia---formerly the world's tallest skyscraper.} \\

\texttt{Entertainment} $\rightarrow$ \texttt{Sport} &
\textit{``So to win something like this really feels great.''} \\

\texttt{Business} $\rightarrow$ \texttt{Tech} &
\textit{It is the second time Vodafone has come under the Commission's scrutiny.} \\
\addlinespace
\multicolumn{2}{c}{\textbf{20 Newsgroups}} \\[2pt]
\texttt{alt.atheism} $\rightarrow$ \texttt{soc.religion.christian} &
\textit{When did I acquire lack of belief in the Easter Bunny?} \\

\texttt{alt.atheism} $\rightarrow$ \texttt{talk.politics.mideast} &
\textit{We liberated it from Spain.} \\

\texttt{rec.sport.baseball} $\rightarrow$ \texttt{sci.med} &
\textit{(Not to mention the possibility that no organ is particularly crucial in some third species.) }\\

\texttt{comp.windows.x} $\rightarrow$ \texttt{soc.religion.christian} &
\textit{Can anyone shoot obvious holes in this theory? }\\

\texttt{alt.atheism} $\rightarrow$ \texttt{soc.religion.christian} &
\textit{I, for one, don't regret it.} \\

\texttt{alt.atheism} $\rightarrow$ \texttt{talk.politics.mideast} &
\textit{(He ignores the challenge that extreemists for peace tend to be quite insistent that everyone accept their ideals for the world, and have even turned quite violent.)} \\

\texttt{rec.autos} $\rightarrow$ \texttt{soc.religion.christian} &
\textit{I won't argue too much about anything.} \\


\texttt{alt.atheism} $\rightarrow$ \texttt{talk.politics.mideast} &
\textit{Their tanks are almost all soviet, with about 100 French tanks (older ones)}. \\

\addlinespace
\multicolumn{2}{c}{\textbf{Wikipedia}} \\[2pt]
\texttt{Language and lit.} $\rightarrow$ \texttt{Music} &
\textit{Tokyopop licensed the series in North America, and released the first two volumes in 2007.} \\

\texttt{Warfare} $\rightarrow$ \texttt{History} &
\textit{Devereux became adjutant to Illinois National Guard Maj. Simon Bolivar Buckner, later a Confederate general during the Civil War and eventually a governor of Kentucky.} \\

\texttt{History} $\rightarrow$ \texttt{Sports and recr.} &
\textit{A friend of English captain Douglas Jardine from their Oxford days, Kent Hughes publicly defended Jardine's tactics of sustained short pitched bowling against the Australian batsmen, arguing that Australia used similar tactics against England during the 1921 Ashes tour.} \\

\texttt{History} $\rightarrow$ \texttt{Language and lit.} &
\textit{The early history of Calakmul is obscure, although a dynastic list has been pieced together that extends back into an ancestral past.} \\


\texttt{Geography} $\rightarrow$ \texttt{Art and arch.} &
\textit{The original house was constructed about 1696 by William Morgan, around the time of his marriage to Margaret Avery.} \\

\texttt{Mathematics} $\rightarrow$ \texttt{Geography} &
\textit{Stevens, Peter S. Patterns in Nature.} \\

\texttt{Language and lit.} $\rightarrow$ \texttt{Sports and recr.} &
\textit{She later became a Distinguished Alumna of the University of North Carolina and the University of Iowa.} \\


\texttt{Sports and recr.} $\rightarrow$ \texttt{History} &
\textit{The delegation was also accompanied by the head boxing coach Lolo Heimuli.} \\

\bottomrule
\end{tabularx}
\label{tab:misassignements}
\end{table}

\subsection{Top-words in ElecDeb60to20}
\label{app:top_words_debate}
\Cref{tab:debate_top_words} lists the twenty topics learned on the ElecDeb60to20 debate between Biden and Trump on 29/10/2020. Most of them correspond to identifiable themes of the discussion, often more specific than the moderator's segments: within the COVID-19 segment, the model separates preventive measures (topic 19) from lockdowns (topic 0); within the candidates' records, it isolates climate and energy (topic 11), forest fires (topic 14) and electric cars (topic 3). A second group of topics captures the register rather than the subject of the debate: conversational interjections (topic 4), personal accusations (topic 1), and claims of achievement (topic 5). These topics are a consequence of modeling individual sentences: in a debate, many sentences carry no policy content, and the model groups them into topics of their own rather than spreading them across the thematic ones. Finally, two topics (9 and 15) contain unrelated words and lack a clear interpretation.
\begin{table}[!t]
\centering
\caption{\textbf{Topics learned on the presidential debate.} For each topic, the ten words with the highest probability under the learned topic--word distribution, and a short label assigned by inspecting them.}
\label{tab:debate_top_words}
\begin{tabularx}{\textwidth}{@{}clX@{}}
\toprule
\textbf{Topic} & \textbf{Label} & \textbf{Top words} \\
\midrule
0  & Lockdowns & \textit{shut, close, country, open, allow, state, democrats, destroy, wrong, throw} \\
1  & Accusations & \textit{man, plan, lie, fraud, office, name, liar, run, wait, acknowledge} \\
2  & Mail-in voting & \textit{ballot, vote, count, send, poll, mail, solicit, state, open, director} \\
3  & Electric cars & \textit{car, cheap, expensive, electric, safe, cost, standard, automobile, tiny, least} \\
4  & Conversational & \textit{guess, wrong, hit, honest, suppose, hey, excuse, surprised, feel, plenty} \\
5  & Achievements & \textit{job, win, top, thank, proud, testify, set, every, happy, administration} \\
6  & Foreign countries & \textit{china, head, die, world, russia, india, eat, billion, ukraine, lunch} \\
7  & Racism & \textit{african, racist, americans, wing, police, community, close, group, racism, whistle} \\
8  & Health care & \textit{drug, care, obamacare, healthcare, affordable, pre, existing, condition, act, million} \\
9  & Mixed & \textit{deal, record, back, god, send, wait, extent, november, soon, period} \\
10 & Economy & \textit{job, economy, create, business, economic, company, history, leave, million, recovery} \\
11 & Climate and energy & \textit{build, plant, deal, paris, accord, energy, emit, hard, environment, problem} \\
12 & Law enforcement & \textit{law, enforcement, police, violence, appropriate, order, crime, support, percent, officer} \\
13 & Election outcome & \textit{election, vote, win, elect, outcome, run, determine, shut, left, party} \\
14 & Forest fires & \textit{forest, tree, management, city, tear, billion, stop, burn, major, else} \\
15 & Mixed & \textit{military, school, bring, support, back, country, church, include, name, class} \\
16 & Taxes & \textit{tax, dollar, billion, eliminate, pay, taxis, million, money, fortune, cut} \\
17 & Deaths and suburbs & \textit{die, suburb, life, place, million, town, home, trouble, far, agree} \\
18 & Supreme Court & \textit{court, justice, hold, leave, supreme, united, states, vote, term, elect} \\
19 & COVID-19 measures & \textit{mask, vaccine, wear, social, speak, test, distancing, end, america, johnson} \\
\bottomrule
\end{tabularx}
\end{table}

\end{document}